\documentclass[10pt,twocolumn,letterpaper]{article}

\usepackage{cvpr}              
\usepackage[table]{xcolor}
\usepackage{multicol} 
\usepackage{balance}

\usepackage{multirow}
\usepackage{tcolorbox}
\usepackage{listings}
\usepackage{xcolor}
\usepackage{caption}

\definecolor{codegray}{RGB}{245,245,245}
\definecolor{codeborder}{RGB}{180,180,180}
\usepackage{booktabs}
\usepackage{array}
\usepackage{multicol}
\usepackage{caption}
\usepackage{xcolor}

\DeclareUnicodeCharacter{00A0}{ }   
\DeclareUnicodeCharacter{200B}{}    

\lstdefinestyle{promptlisting}{
  basicstyle=\ttfamily\tiny, 
  breaklines=true,
  showstringspaces=false,
  frame=none,
  columns=fullflexible,
  backgroundcolor=\color{codegray},
  tabsize=2,
  xleftmargin=0pt,
  xrightmargin=0pt,
  aboveskip=0pt,
  belowskip=0pt,
  postbreak=\mbox{}, breakindent=0pt,
}
\usepackage{graphicx}
\usepackage{booktabs}
\usepackage{tabularx}
\newcolumntype{Y}{>{\raggedright\arraybackslash}X}
\usepackage{longtable}
\usepackage{adjustbox}
\usepackage{lipsum}

\usepackage{tabularx}
\newcolumntype{Y}{>{\raggedright\arraybackslash}X}
\newcolumntype{R}{>{\raggedright\arraybackslash}X}
\newcolumntype{P}{>{\raggedright\arraybackslash}X}
\newcolumntype{S}[1]{>{\raggedright\arraybackslash\hsize=#1\hsize}X}
\usepackage{pifont}

\definecolor{cvprblue}{rgb}{0.21,0.49,0.74}
\usepackage[pagebackref,breaklinks,colorlinks,allcolors=cvprblue]{hyperref}

\def\paperID{17614} 
\def\confName{CVPR}
\def\confYear{2026}

\title{Pixels Don't Lie (But Your Detector Might): Bootstrapping MLLM-as-a-Judge for Trustworthy Deepfake Detection and Reasoning Supervision }

\author{Kartik Kuckreja$^1$     Parul Gupta$^2$ Muhammad Haris Khan$^1$ Abhinav Dhall$^2$\\
$^1$MBZUAI \quad 
$^2$Monash University  \thanks{
{\tt \{kartik.kuckreja,muhammad.haris\}@mbzuai.ac.ae} \\
  {\tt \{parul,abhinav.dhall\}@monash.edu}\\}
}

\begin{document}
\maketitle

\begin{abstract}
Deepfake detection models often generate natural-language explanations, yet their reasoning is frequently ungrounded in visual evidence, limiting reliability. Existing evaluations measure classification accuracy but overlook reasoning fidelity. We propose DeepfakeJudge, a framework for scalable reasoning supervision and evaluation, that integrates an out-of-distribution benchmark containing recent generative and editing forgeries, a human-annotated subset with visual reasoning labels, and a suite of evaluation models, that specialize in evaluating reasoning rationales without the need for explicit ground truth reasoning rationales. The Judge is optimized through a bootstrapped generator–evaluator process that scales human feedback into structured reasoning supervision and supports both pointwise and pairwise evaluation. On the proposed meta-evaluation benchmark, our reasoning-bootstrapped model achieves an accuracy of 96.2\%, outperforming \texttt{30x} larger baselines. The reasoning judge attains very high correlation with human ratings and 98.9\% percent pairwise agreement on the human-annotated meta-evaluation subset. These results establish reasoning fidelity as a quantifiable dimension of deepfake detection and demonstrate scalable supervision for interpretable deepfake reasoning. Our user study shows that participants preferred the reasonings generated by our framework 70\% of the time, in terms of faithfulness, groundedness, and usefulness, compared to those produced by other models and datasets.  All of our datasets, models, and codebase are \href{https://github.com/KjAeRsTuIsK/DeepfakeJudge}{open-sourced}.
\end{abstract}

\section{Introduction}
\label{sec:intro}

Recent advances in diffusion and transformer-based models such as \textbf{Stable Diffusion}~\cite{rombach2022highresolutionimagesynthesislatent}, \textbf{DALL·E 2}~\cite{ramesh2022hierarchicaltextconditionalimagegeneration}, and \textbf{Imagen}~\cite{saharia2022photorealistictexttoimagediffusionmodels} have made synthetic images nearly indistinguishable from real ones, posing new challenges for visual forensics. Early detectors relied on low-level cues like frequency inconsistencies~\cite{durall2020watchupconvolutioncnnbased} or eye-blinking patterns~\cite{li2018ictuoculiexposingai}, but these fail under modern generation pipelines. Deep learning-based detectors~\cite{nguyen2018capsuleforensicsusingcapsulenetworks, jiang2020deeperforensics10largescaledatasetrealworld, rössler2019faceforensicslearningdetectmanipulated} improve accuracy but generalize poorly to out-of-distribution (OOD) data, evaluating unseen models critical for realistic benchmarking.

\begin{figure}[t]
    \centering
    \includegraphics[width=\columnwidth]{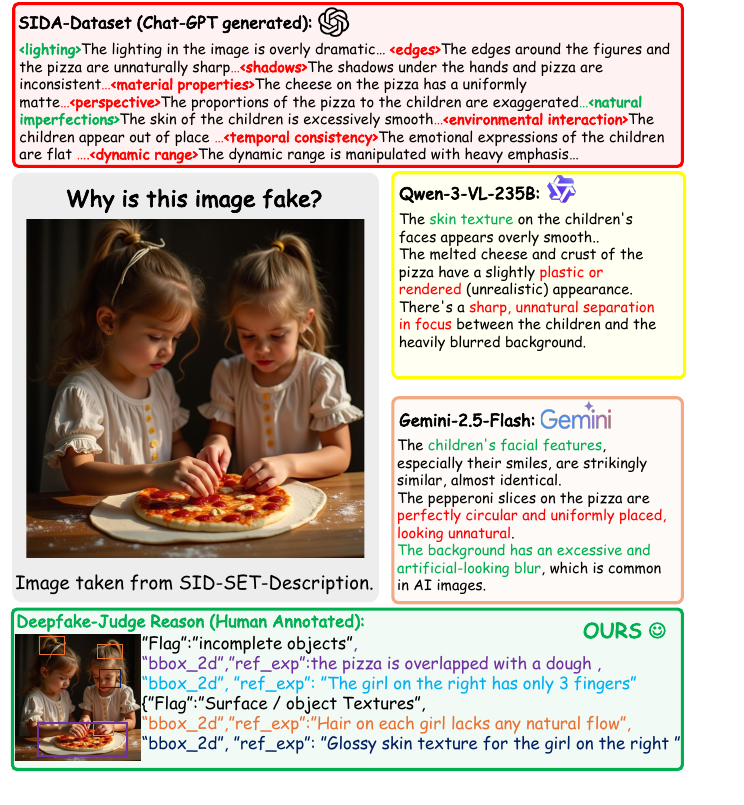}
    \caption{Comparison of reasoning rationales from SIDA, Qwen-3-VL-235B, Gemini-2.5-Flash, and Human Annotation in our proposed DeepfakeJudge-Reason. \textcolor{red}{Red} and \textcolor{green}{green} indicate incorrect and correct flags respectively. Our human annotations provide dense, localized, and accurate reasoning.}
    \label{fig:teaser_reasoning}
\end{figure}

Beyond classification accuracy, reliable deepfake detection requires interpretable and visually grounded reasoning. Recent explainable detectors such as \textbf{SIDA}~\cite{huang2025sidasocialmediaimage}, \textbf{GenBuster}~\cite{wen2025busterxunifiedcrossmodalaigenerated}, and \textbf{FakeShield}~\cite{xu2025fakeshieldexplainableimageforgery} attempt textual explanations, but these are often ungrounded or hallucinated, similar to issues seen in large vision-language models (VLMs)~\cite{liu2024mitigatinghallucinationlargemultimodal, guan2024hallusionbenchadvanceddiagnosticsuite}. As shown in Figure~\ref{fig:teaser_reasoning}, our analysis finds that existing systems frequently misattribute manipulations, describing lighting or color inconsistencies instead of real artifacts in geometry, texture, or physical plausibility. Prior works~\cite{vo2025visionlanguagemodelsbiased, liu2024reducinghallucinationsvisionlanguagemodels} show that VLMs over-rely on textual priors and world knowledge, often ignoring visual cues; for eg., predicting that a cow has four legs even when an image shows three. Consequently, VLM-based approaches ~\cite{huang2025sidasocialmediaimage} tend to produce ungrounded and unreliable rationales.

\begin{figure}[t]
    \centering
    \includegraphics[width=\columnwidth]{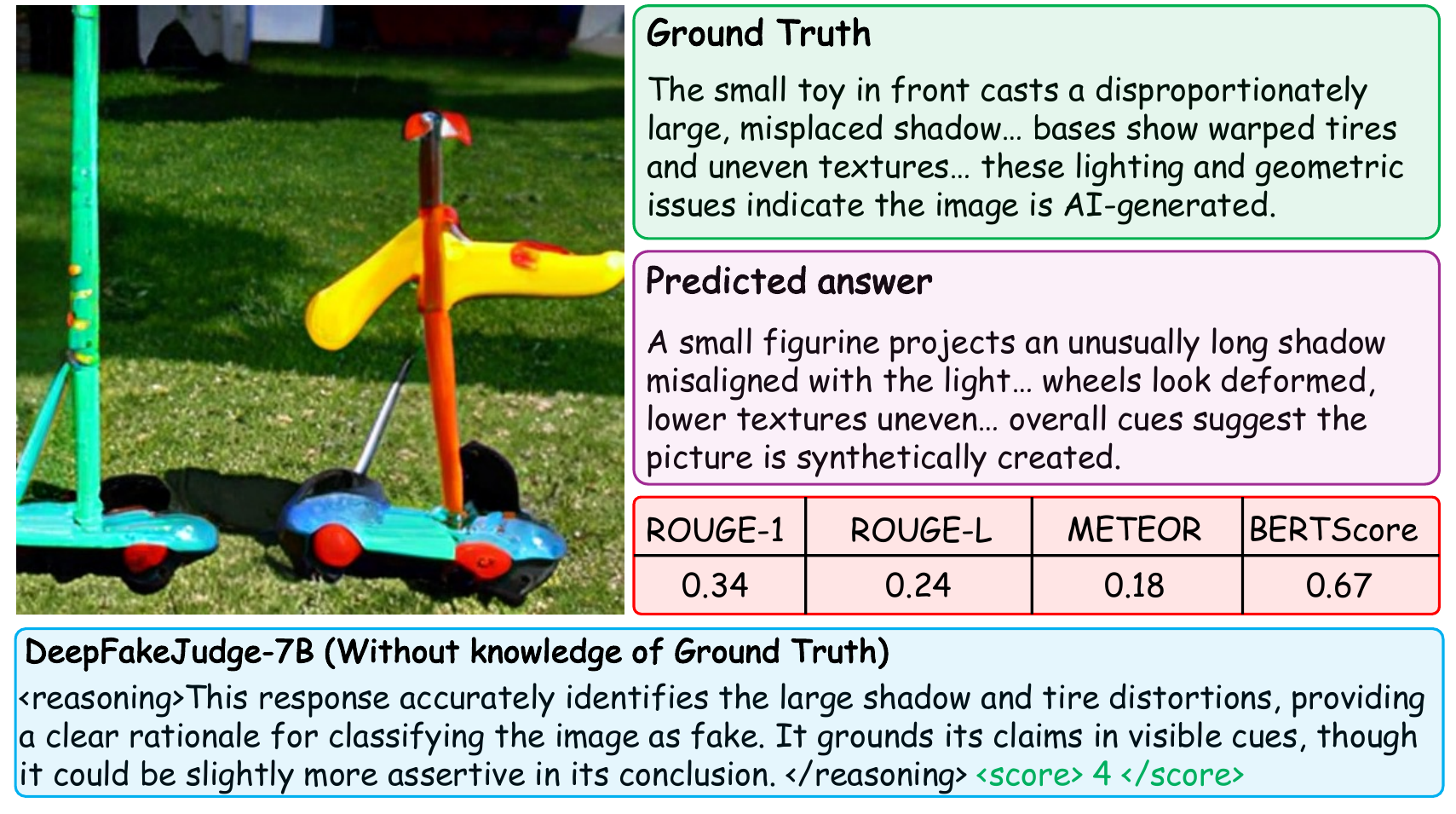}
    \caption{Existing metrics (ROUGE, METEOR, BERTScore) fail to capture reasoning quality. DeepFakeJudge directly evaluates the image, providing both a reasoning quality score and a rationale.}
    \label{fig:metric_justify}
\end{figure}

Traditional text metrics such as BLEU~\cite{papineni-etal-2002-bleu}, ROUGE~\cite{lin-2004-rouge}, and METEOR\cite{banerjee-lavie-2005-meteor}, as well as semantic ones like BERTScore~\cite{zhang2020bertscoreevaluatingtextgeneration}, capture linguistic overlap but not factual grounding, correlating poorly with human judgment~\cite{zheng2023judgingllmasajudgemtbenchchatbot}. Our qualitative experiments confirm these limitations (Figure~\ref{fig:metric_justify}), underscoring the need for a multimodal evaluation framework that measures reasoning fidelity and visual grounding.

\begin{figure*}[t]
    \centering
    \includegraphics[width=\textwidth]{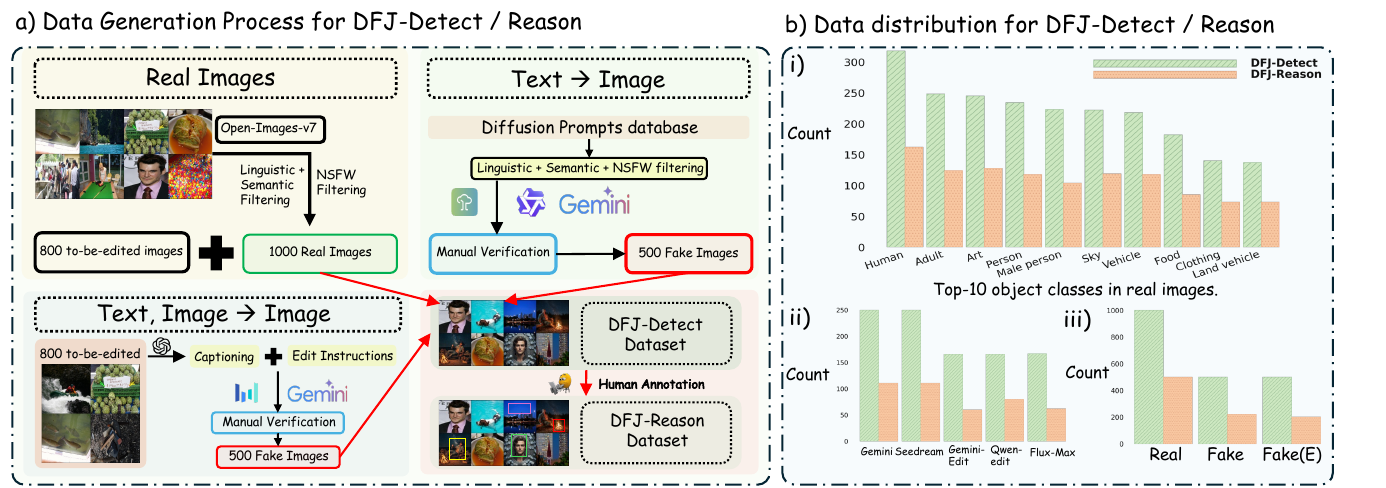}
    \caption{a) Data generation process for OOD-dataset generation b) Data Distribution for DeepfakeJudge-Detect/Reason splits. i) Shows the distribution of top-10 classes in the real-subset, ii) shows the distribution of generation models, and iii) shows the label-wise distribution. }\vspace{-1mm}
    \label{fig:ood_dataset}
\end{figure*}
To address these issues, we introduce a unified framework for reasoning supervision and evaluation in deepfake detection. In particular, we propose a comprehensive benchmark: \textbf{DeepfakeJudge} comprising of different components, which evaluate the OOD (Out-of-Distribution) (\textit{DeepfakeJudge-Detect} and \textit{DeepfakeJudge-Reason}), in-distribution (\textit{DeepfakeJudge-Meta} and \textit{DeepfakeJudge-Meta-Human}), detection and reasoning capabilities of deepfake detection methods. \textbf{Our main contributions are:}
\begin{itemize}
    \item \textbf{OOD Deepfake Benchmark:} We combine text-to-image and editing-based models with real images to test both detection and reasoning generalization, comparing over 15 State-of-the-Art VLMs.
    \item \textbf{Human-Annotated Reasoning Set:} We attach textual explanations to spatial evidence through bounding boxes and referring expressions, densely annotated by humans for fine-grained reasoning supervision.
    \item \textbf{Novel Bootstrapped Supervision Process:} We scale human annotations into structured reasoning–rating data using an iterative generator–evaluator pipeline.
    \item \textbf{VLM-Based Reasoning Judge:} A multimodal evaluator that assesses explanations through pointwise and pairwise comparisons aligned with human preferences. It provides both a rating and a concise rationale, serving as a new metric for measuring reasoning quality directly from the image, without requiring explicit ground-truth reasoning. 
\end{itemize}

Together, these components establish a unified foundation for evaluating and improving reasoning fidelity in deepfake detection, addressing both generalization and interpretability in modern forensic systems.

\section{Related Work}

\noindent\textbf{Deepfake Detection.}
Deepfake detection research has progressed rapidly from low-level forensic analysis to multimodal understanding. Early approaches relied on handcrafted or signal-level features such as frequency inconsistencies~\cite{durall2020watchupconvolutioncnnbased}, blending boundaries~\cite{rössler2019faceforensicslearningdetectmanipulated}, or blink patterns~\cite{li2018ictuoculiexposingai} to detect visual anomalies in manipulated faces. With the advent of deep generative models, learning-based methods became dominant, leveraging convolutional or recurrent architectures~\cite{nguyen2018capsuleforensicsusingcapsulenetworks, jiang2020deeperforensics10largescaledatasetrealworld, rössler2019faceforensicslearningdetectmanipulated} to capture spatial and temporal cues. Later works explored attention mechanisms and relational reasoning~\cite{ li2020facexraygeneralface}, and several large-scale benchmarks such as FaceForensics++~\cite{rössler2019faceforensicslearningdetectmanipulated}, DFDC~\cite{dolhansky2020deepfakedetectionchallengedfdc}, and DeeperForensics~\cite{jiang2020deeperforensics10largescaledatasetrealworld} established standard evaluation protocols. 
More recently, transformer-based and diffusion-oriented detectors have been introduced to improve generalization across generation pipelines~\cite{choi2024exploitingstylelatentflows, yan2024transcendingforgeryspecificitylatent, tan2023rethinkingupsamplingoperationscnnbased}. Together, these studies have shaped a rich landscape of deepfake detection approaches spanning signal processing, deep representation learning, and multimodal fusion.

\noindent\textbf{Reasoning and Explainability.}
Explainability has become an important direction in AI-based visual forensics, driven by the need for interpretable and trustworthy predictions. Early explainable methods in computer vision focused on saliency and attention maps~\cite{simonyan2014deepinsideconvolutionalnetworks, Selvaraju_2019}, while later works incorporated explicit reasoning modules to produce human-understandable justifications~\cite{arrieta2019explainableartificialintelligencexai, doshivelez2017rigorousscienceinterpretablemachine}. In deepfake detection, several vision–language approaches have been proposed to couple classification with textual reasoning. Models such as SIDA~\cite{huang2025sidasocialmediaimage}, GenBuster++~\cite{wen2025busterxunifiedcrossmodalaigenerated}, and FakeShield~\cite{xu2025fakeshieldexplainableimageforgery} extend detection frameworks with generative explanation capabilities, but they suffer from the same set of issues in reasoning, over-reliance on text inputs, and non-grounded reasoning traces. Related studies in human-centered explainable AI~\cite{Zhang_2022} further highlight the role of reasoning in improving transparency and trust between humans and AI systems.

\noindent\textbf{Large Language Models as Evaluators.}
Large language models have recently been used as evaluators for assessing text quality and reasoning in natural language tasks~\cite{ zheng2023judgingllmasajudgemtbenchchatbot,lambert2024rewardbenchevaluatingrewardmodels}. This has inspired the \textit{LLM-as-a-Judge} paradigm, in which models act as automated evaluators of response quality or factual correctness. In the multimodal domain, recent benchmarks such as MME~\cite{fu2025mmecomprehensiveevaluationbenchmark}, and SEED-Bench 2~\cite{li2023seedbench2benchmarkingmultimodallarge} extend this idea to vision–language reasoning, evaluating alignment between visual inputs and textual outputs. Parallel research has explored multimodal hallucination and evaluation frameworks~\cite{bang2025hallulensllmhallucinationbenchmark, guan2024hallusionbenchadvanceddiagnosticsuite}, paving the way for using vision-language models as scalable judges of explanation quality. 

These collective developments across detection, reasoning, and multimodal evaluation naturally motivate our work, \textbf{\textit{DeepFakeJudge}}, which unifies these directions into a comprehensive framework for reasoning supervision and assessment to complement deepfake detection.

\begin{figure*}[t]
    \centering
    \includegraphics[width=\textwidth]{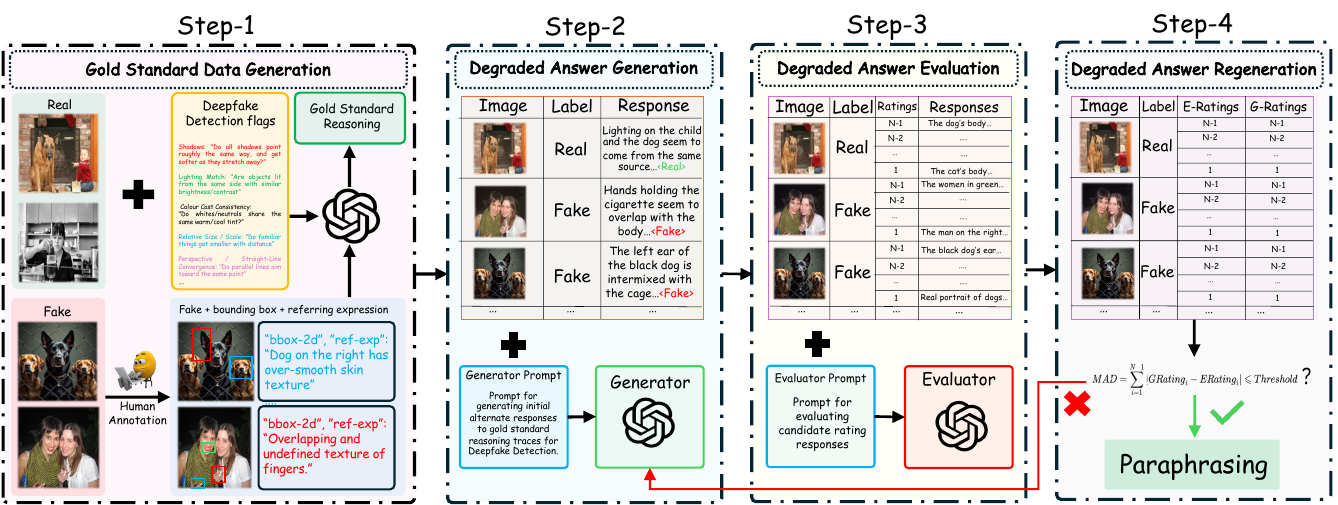}
    \caption{Overview of the DeepFakeJudge bootstrapping pipeline. 
    {Step 1:} Generating gold standard reasoning rationales using the in-domain human annotated dataset (Section \ref{subsec:human-annotation}) {Step 2: }The generator creates reasoning responses for each image–label pair across five rating levels. 
    {Step 3: }The evaluator provides feedback and re-scores the responses until alignment is achieved. 
    {Step 4: }All accepted responses are paraphrased to create stylistically diverse but semantically consistent data.}
    \label{fig:bootstrap_framework}
\end{figure*}

\section{Methodology}

We pursue scalable evaluation of reasoning quality in deepfake detection through a VLM-based judge trained on human-annotated data. Our framework has the following three stages: (1) Dataset construction (Section~\ref{subsection:OOD_dataset_construction}), (2) Bootstrapping human annotation for scalable reasoning (Section~\ref{subsection:bootstrapping_human}), and (3) DeepFakeJudge training (Section~\ref{subsubsection:judge_training}). Figure \ref{fig:ood_dataset} refers to stage 1 of our framework, while Figure \ref{fig:bootstrap_framework} refers to stage 2.


\subsection{Dataset Construction}
\label{subsection:OOD_dataset_construction}

\noindent\textbf{Real Image Collection.}\label{real_image_collection}
To construct the real subset of our Out-of-Distribution (OOD) dataset, we build upon the official Open-Images~V7 \cite{OpenImages} dataset. 
We first get class descriptions, image-level labels, and bounding-box annotations, ensuring full consistency with Google’s schema. 
Secondly, we then merge human-verified image-level labels with bounding boxes to form a pool of densely annotated candidate images, and then apply a 
a seeded stochastic greedy set-cover algorithm to maximize label diversity while maintaining reproducibility. 
From this pool, we select $1{,}000$ diverse candidates for our real subset, and an additional $800$ images are reserved for the \emph{editing} subset. Figure \ref{fig:ood_dataset} b) i) shows the representations of the top-10 classes present in the real subset.  

\noindent\textbf{Fake Data Generation.}
We generate our fake data using text-to-image and text+image-to-image(editing), with generation models selected from the Image-to-Text and Image-Editing Leaderboards from Artificial-Analysis \cite{artificialanalysis2025}.

\noindent\textbf{Text \texttt{-->} Image.}
To construct the T2I (Text-to-Image) subset, we extracted realistic, photography-oriented prompts from a large-scale diffusion prompt dataset \cite{wangDiffusionDBLargescalePrompt2022}. 
Prompts were filtered using a series of linguistic and semantic heuristics designed to retain only English, non-fantasy, non-NSFW descriptions emphasizing real-world photographic content. 
Each candidate was scored for textual richness and dense grounding, then categorized into broad content types such as portraits, landscapes, and objects (see Appendix Section \ref{tab:prompt-filtering} for details about the filtering process). 
A balanced selection procedure ensured diversity across these categories, resulting in a pool of $2{,}000$ high-quality prompts.
These prompts were then refined through \texttt{GPT\text{-}4o\text{-}mini} \cite{openai_gpt4o_mini_2024} to align stylistically with the input format expected by \texttt{Gemini} \cite{comanici2025gemini25pushingfrontier} and \texttt{SeedDream} \cite{seedream2025seedream40nextgenerationmultimodal}. 
We used both \texttt{Gemini} \cite{comanici2025gemini25pushingfrontier} and \texttt{SeedDream} \cite{seedream2025seedream40nextgenerationmultimodal}; each generating half of the samples to synthesize the corresponding images. 
From the generated outputs, $700$ images were randomly sampled and further manually inspected for realism and fidelity, yielding the final curated set of $500$ fake images.

\noindent\textbf{Text, Image \texttt{-->} Image (editing).}
For the TI2I (Text+Image-to-Image) subset, we used the $800$ real images described in Section~\ref{real_image_collection}. 
Ground-truth captions were first generated for each image using \texttt{GPT\text{-}4o\text{-}mini} \cite{openai_gpt4o_mini_2024}. 
Based on both the image and its caption, we prompt the model to then produced three candidate edit instructions per sample.
One edit instruction is randomly selected for each image and applied independently by any one of the three text-to-image editing models: \texttt{Gemini-Nano Banana} \cite{comanici2025gemini25pushingfrontier}, \texttt{Flux-Kontext-Max} \cite{labs2025flux1kontextflowmatching}, and \texttt{Qwen-Edit-2509} \cite{wu2025qwenimagetechnicalreport}. 
Each model generated an equal number of edited outputs, yielding the second chunk of 500 fake images. Figure \ref{fig:ood_dataset} b) ii) and iii) show the data distribution of real, fake, and edited images in DeepfakeJudge-Detect/Reason datasets.

\noindent\textbf{Human Annotation for Reasoning}
\label{subsec:human-annotation}
We conduct a detailed human annotation process to generate accurate and interpretable reasoning rationales. Six trained annotators label 1,500 fake images drawn from both in-distribution and OOD datasets, which are further manually checked for verification. The in-distribution set includes samples from MultifakeVerse \cite{gupta2025multiversedeepfakesmultifakeversedataset}, SID-SET-Description \cite{huang2025sidasocialmediaimage}, and Community-Forensics \cite{park2025communityforensicsusingthousands}, with a total of 1025 samples (split between real / fake / edited) and is used to train our DeepfakeJudge models. The OOD set, DeepfakeJudge-Reason, contains 924 samples (500 real + 424 fake) randomly selected from DeepfakeJudge-Detect dataset.

\begin{figure*}[b]
    \centering
    \includegraphics[width=\textwidth]{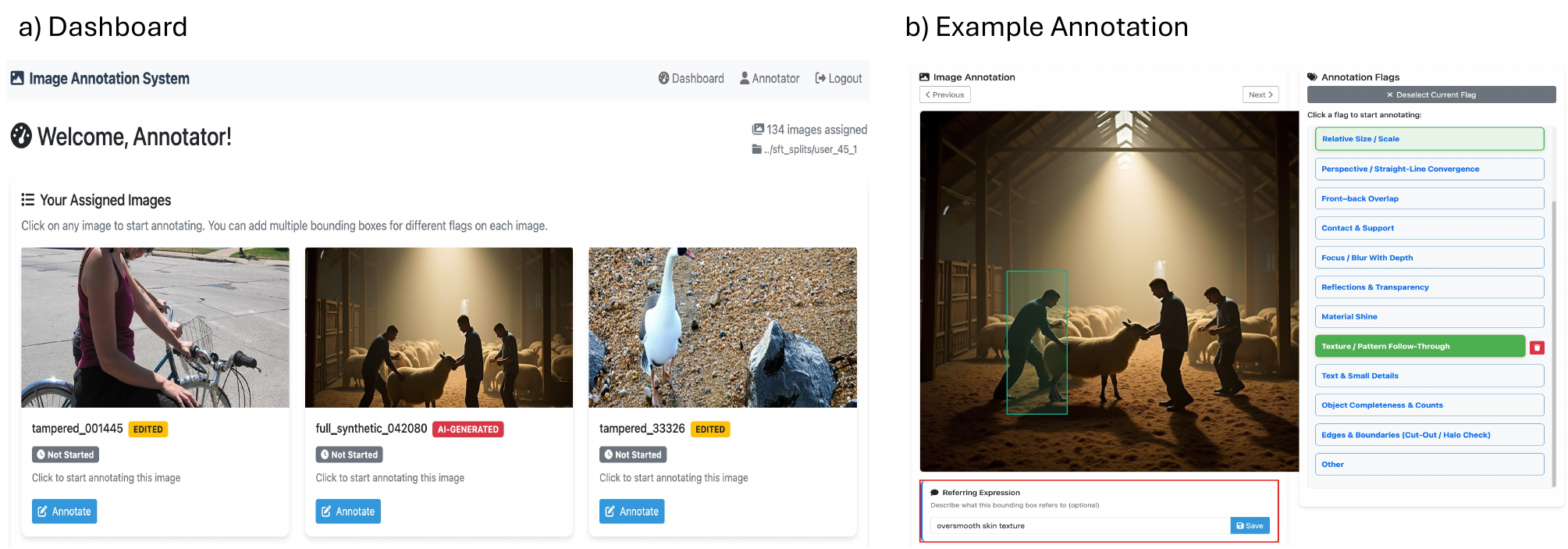}
    \caption{Overview of the human annotation framework. a) Shows the user dashboard, b) shows an example of the annotation process. The user selects a flag, draws the corresponding bounding box over the image, and then writes a referring expression.}
    \label{fig:annotation_framework}
\end{figure*}

\noindent Annotators view each image with its ground-truth label (fake or edited), and for edited images, the corresponding generation instruction. They select relevant visual artifact flags (see Appendix Table \ref{tab:flag-taxonomy}), draw bounding boxes around affected regions, and add short descriptions of the anomalies (Appendix Figure \ref{fig:annotation_framework}). To ensure consistency, all annotators complete 10 shared pilot samples before working independently on disjoint subsets. Inter-annotator agreement, measured by Cohen’s $\kappa=0.71$, indicates substantial alignment. Finally, each annotated sample (image, label, and annotations) is processed by GPT-4o-mini \cite{openai_gpt4o_mini_2024} to produce gold-standard reasoning rationales (Prompt is shown in Appendix).

\subsection{Bootstrapping Human Annotation for Scalable Reasoning Supervision}
\label{subsection:bootstrapping_human}

Our bootstrapping pipeline uses a generator–evaluator framework to create and validate reasoning data for image authenticity assessment. It builds on VideoJudge \cite{waheed2025videojudgebootstrappingenablesscalable} and prior work on self-consistency and self-improvement in large language models \cite{mitchell2022enhancingselfconsistencyperformancepretrained,wang2023selfconsistencyimproveschainthought,chen2023universalselfconsistencylargelanguage,weng-etal-2023-large}.
The process has two main stages: (1) iterative bootstrapping to build large-scale, fine-grained reasoning and rating data (section \ref{sec:bootstrap_start}), and (2) fine-tuning (section \ref{subsubsection:judge_training}) a Vision-Language model for pointwise and pairwise scoring (see Figure~\ref{fig:bootstrap_framework}). For each image $I$ with ground-truth label $g$, the generator $G$ produces reasoning responses across rating levels $r \in {1,\ldots,5}$. The gold reasoning $y^{}$, obtained from GPT for real images and from human annotations for fake ones, represents the highest-quality reference. The evaluator $E$ then scores each generated rationale, keeping only those where the predicted and target ratings match. Mismatched cases are refined through feedback, and all accepted reasonings (including $y^{}$) are paraphrased to reduce stylistic bias. The final graded corpus is used to train our DeepfakeJudge models.

\begin{figure}[htbp]
    \centering
    \includegraphics[width=\columnwidth]{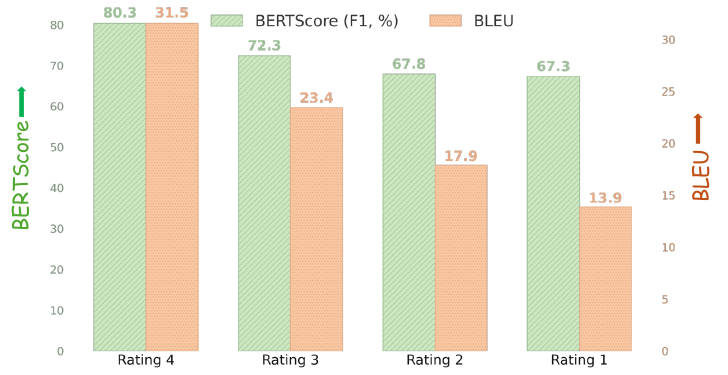}
    \caption{Comparison of BERTScore and BLEU of candidate ratings against the gold standard ratings. Our data bootstrapping method generates reasonings of continuously degrading quality.}
    \label{fig:bleu_meteor}
\end{figure}


\subsubsection{Bootstrapping Process}

We start from the in-distribution annotated dataset described in Section~\ref{subsec:human-annotation}, which consists of image–label pairs and their associated gold reasoning responses. Each sample is represented as a triplet $(I, g, y^{*})$, where $I$ is the image, $g$ the ground-truth authenticity label, and $y^{*}$ the high-quality reasoning response.

\noindent\textbf{Initial Generation:}\label{sec:bootstrap_start}
For each $(I, g, y^{*})$, the generator produces $(N-1)$ reasoning responses, one per rating level $r \in \{1, \ldots, N-1\}$.  
The generation step is formalized as:
\begin{equation}
    y^{(r)}_{0} = G(p_{\text{gen}} \, \| \, I \, \| \, g \, \| \, y^{*}, r),
\end{equation}
where $p_{\text{gen}}$ is the generation prompt that conditions on image, ground-truth label, gold reasoning, and intended rating.

\noindent\textbf{Feedback and Evaluation:}
Each generated reasoning $y^{(r)}_{t}$ is assessed by an evaluator $E$, which outputs a predicted rating $\hat{r}$ and a feedback rationale $f^{(r)}_{t}$.  
The evaluation process is expressed as:
\begin{equation}
    \hat{r}, f^{(r)}_{t} = E(p_{\text{eval}} \, \| \, I \, \| \, g \, \| \, y^{*} \, \| \, y^{(r)}_{t}),
\end{equation}
and the rating deviation is computed as:
\begin{equation}
    \Delta^{(r)}_{t} = |r - \hat{r}|.
\end{equation}
Candidates that satisfy $\Delta^{(r)}_{t} \leq \alpha$ are directly accepted; otherwise, they are refined.

\noindent\textbf{Refinement:}
For candidates with a rating deviation greater than $\alpha$, the generator is re-prompted using evaluator feedback. The refinement process continues until the candidate meets the acceptance threshold or reaches a maximum number of iterations $T$:
\begin{equation}
    y^{(r)}_{t+1} = G(p_{\text{ref}} \, \| \, I \, \| \, g \, \| \, y^{*} \, \| \, y^{(r)}_{t} \, \| \, f^{(r)}_{t}, r).
\end{equation}
A reasoning response $y^{(r)}_{t}$ is added to the bootstrapped dataset if $|r - \hat{r}| \leq \alpha$.

\noindent\textbf{Paraphrasing Step:}
Once all rating levels have aligned responses, each reasoning, including the gold one, is paraphrased five times to generate stylistically diverse variants.  
This step ensures that the evaluator focuses on the semantic and logical consistency of reasoning rather than memorizing linguistic patterns. Section \ref{section: paraphrase_qualitative} in Appendix shows qualitative samples of paraphrased gold-standard ratings.

To evaluate whether our paraphrases maintain semantic similarity while introducing lexical variation, we compute BERTScore and BLEU on a random sample of 500 pairs of original and paraphrased ratings. BERTScore measures semantic similarity (with 1 indicating perfect similarity and 0 indicating none), while BLEU measures lexical overlap (with 1 indicating identical wording and 0 indicating no overlap). We obtain an average \textbf{BERTScore of 0.92 and a BLEU score of 0.39}, suggesting that the paraphrases preserve meaning while differing in surface form.

\begin{figure}[t]
    \centering
    \includegraphics[width=\columnwidth]{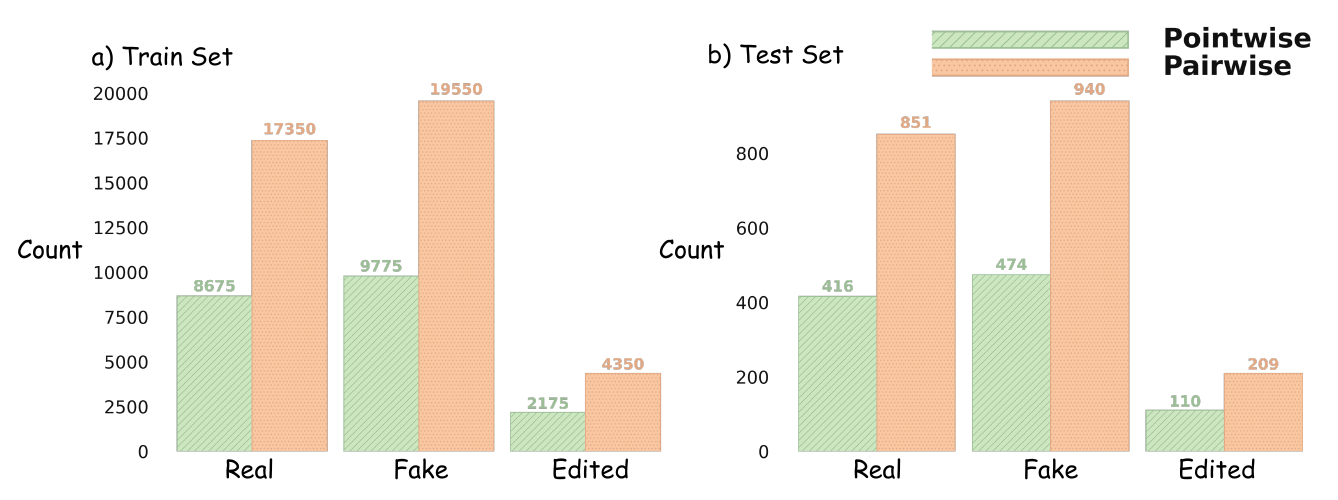}
    \caption{Data distribution for DeepfakeJudge-Meta dataset.}
    \label{fig:dfj_data_split}
\end{figure}
\vspace{-0.5mm}

\noindent\textbf{DeepfakeJudge-Meta:}
The final bootstrapped dataset is composed of tuples $\{(I, g, y, r)\}$, where each image–label pair is associated with five rating levels and multiple paraphrased reasoning samples per level. The total dataset distribution is shown in the figure \ref{fig:dfj_data_split}. To verify whether the ratings are actually degraded, we calculate BLEU \cite{papineni-etal-2002-bleu} and BERTScore \cite{zhang2020bertscoreevaluatingtextgeneration} and average the scores for each rating class. Figure \ref{fig:bleu_meteor} verifies our claim that our pipeline indeed produces linearly degrading ratings. 
This curated dataset provides the foundation for training both pointwise and pairwise evaluator models. Qualitative examples of our gold standard reasoning, along with degraded quality ratings is available in Appendix Section \ref{sec: qualitative_examples}.



\begin{table*}[t]
\centering
\small
\caption{Comparison of SOTA open-sourced, closed-source, reasoning, and deepfake models on the DeepfakeJudge-Detect datasets.}
\begin{tabular}{l lcccccc}
\toprule
\textit{Type} & \textit{Model} & \textit{Real Acc} & \textit{Real F1} & \textit{Fake Acc} & \textit{Fake F1} & \textit{Overall Acc} & \textit{Overall F1} \\
\midrule
\multirow{2}{*}{Closed} 
 & Gemini-2.5-Flash & 96.6 & 73.7 & 34.5 & 50.0 & 65.5 & 61.9 \\
 & ChatGPT-4o-mini & 95.8 & 70.2 & 22.7 & 35.8 & 59.3 & 53.0 \\
\midrule
\multirow{6}{*}{Open} 
 & InternVL3.5-1B-HF & 47.8 & 63.0 & 47.8 & 11.4 & 47.8 & 37.2 \\
 & Qwen3-VL-2B-Instruct & 49.8 & 44.7 & 49.8 & 54.0 & 49.8 & 49.3 \\
 & Qwen-3-VL-8B-Instruct & 50.4 & 23.7 & 50.4 & 63.2 & 50.4 & 43.5 \\
 & Google-Gemma-12B & 57.7 & 57.4 & 49.4 & 54.0 & 66.1 & 60.8 \\
 & Microsoft-Phi-4-Instruct & 61.0 & 60.8 & 54.4 & 58.2 & 67.5 & 63.4 \\
 & InternVL3.5-GPT-OSS-20B-A4B & 55.6 & 67.6 & 55.6 & 29.2 & 55.6 & 48.4 \\
 & Qwen-3-VL-30B & 94.6 & 74.5 & 41.0 & 56.0 & 67.7 & 65.3 \\
 & Qwen-3-VL-235B & 93.5 & 78.6 & 55.4 & 68.4 & \textbf{74.5} & \textbf{73.5} \\
\midrule
\multirow{3}{*}{Reasoning} 
 & Qwen-3-VL-8B-Thinking & 67.1 & 67.1 & 78.7 & 69.9 & 55.5 & 64.3 \\
 & Qwen-3-VL-30B-Thinking & 67.4 & 66.0 & 87.6 & 72.9 & 47.2 & 59.1 \\
 & Qwen-3-VL-235B-Thinking & 75.0 & 76.6 & 90.3 & 79.8 & 63.7 & 73.4 \\
\midrule
\multirow{2}{*}{DF} 
 & SIDA-13B-Description & 67.6 & 57.0 & 27.9 & 34.5 & 48.1 & 45.8 \\
 & Qwen2.5-VL-Gen-Buster++ & 49.9 & 40.0 & 49.9 & 66.5 & 49.9 & 33.5 \\
\bottomrule
\end{tabular}
\label{tab:detect_revised}
\end{table*}

\noindent\textbf{DeepfakeJudge-Meta-Pointwise:} 
In the pointwise setting, the model receives a tuple consisting of the input image, the ground-truth authenticity label, and a candidate reasoning response. The task is to predict a rating between 1 and 5 enclosed within \texttt{<score>...</score>}, along with a short justification enclosed within \texttt{<reasoning>...</reasoning>}. This setting evaluates the model’s ability to assign an absolute quality score to a single reasoning trace. Using our dataset, we construct 20,625 image–label–response tuples for training and 1,000 for testing, forming the \textit{DeepfakeJudge-Meta-Pointwise} subset.

\noindent\textbf{DeepfakeJudge-Meta-Pairwise:} 
In the pairwise setting, the model is provided with an input image, its ground-truth label, and two candidate reasoning responses. The model must decide which response presents a stronger, more grounded rationale, outputting its choice as \texttt{<answer>A or B</answer>}. The order of options is randomized to avoid positional bias. From our dataset, we create 41,250 image–label–response pairs for training and 2,000 for testing, forming the \textit{DeepfakeJudge-Meta-Pairwise} subset.

\noindent\textbf{DeepfakeJudge-Meta-Human:}
 To verify the consistency between model predictions and human reasoning judgments, we conduct a human annotation study for both pointwise and pairwise evaluation tasks. Two expert annotators independently labeled 100 overlapping samples for each task, allowing us to measure inter-annotator agreement on reasoning quality. After the initial annotation, we filter samples in which both annotators agree on the same rating. This final subset is referred to as \textit{DeepfakeJudge-Human}.
Across both evaluation tasks, the annotators exhibit strong consistency, with a raw agreement of 0.90 and a Cohen’s $\kappa \approx 0.80$ for the pairwise evaluation, and a mean MSE of $0.39$ for pointwise, indicating strong coherence and inter-annotator reliability in ratings. These results confirm the quality and consistency of human reasoning supervision in our evaluation framework. Appendix(\ref{sec:agreement_stats}) show all the prompts and the inter-annotator agreement statistics.

\subsubsection{DeepFakeJudge Training}
\label{subsubsection:judge_training}

We train our evaluator models using the bootstrapped dataset $D = \{(I_i, g_i, y_i, t_i)\}_{i=1}^{M}$, where $I_i$ denotes the image, $g_i$ its ground-truth label, $y_i$ a reasoning response (or a reasoning pair in the pairwise setting), and $t_i$ the target output, such as a numerical rating or a preference label.  
The evaluator model $E_{\theta}$ is optimized via the standard negative log-likelihood objective:
\begin{equation}
    \mathcal{L}(\theta) = -\frac{1}{M} \sum_{i=1}^{M} \sum_{j=1}^{|t_i|} 
    \log P_{\theta}(t_{i,j} \mid t_{i,<j}, I_i, g_i, y_i),
\end{equation}
where $t_{i,j}$ is the $j$-th token of the target sequence.
We train 2 models, \texttt{Qwen-2.5-VL-7B} \cite{bai2025qwen25vltechnicalreport} and \texttt{Qwen-2.5-VL-3B} \cite{bai2025qwen25vltechnicalreport} for both, pairwise and pointwise settings. We train each model for 2 epochs, on  $20,625$ samples for pointwise, and  $20,625$ samples (randomly sampled from $41,250$ total samples) for pairwise evaluation. Detailed training hyperparameters are available are shown in Appendix (Section \ref{sec:training_hyperparameters}).

\begin{table*}[t]
\centering
\caption{Evaluation of models on the DeepfakeJudge-Reason. BLEU ($\text{B-1–3}$), ROUGE ($\text{R-1–L}$), METEOR, and BERTScore are normalized to [0, 1]. DeepfakeJudge-3B (DFJ-3B) provides more consistent comparative ratings on a 1–5 scale, whereas other metrics yield inconclusive results. For all of the above metrics, higher score means better performance.}
\small
\begin{tabular}{l lccccccccc}
\toprule
{\textbf{Type}} & {\textbf{Model}}& {\textbf{B-1}}& {\textbf{B-2}}& {\textbf{B-3}}& {\textbf{R-1}} & {\textbf{R-2}} & {\textbf{R-L}} & {\textbf{METEOR}} & {\textbf{BERT}} & {\textbf{DFJ-3B}}($\uparrow$) \\
\midrule
\multirow{2}{*}{{Closed}} 
 & Gemini-2.5-Flash & 0.05 & 0.02 & 0.02 & 0.30 & 0.05 & 0.17 & 0.17 & 0.60 & 3.17 \\
 & ChatGPT-4o-mini & 0.01 & 0.01 & 0.01 & 0.15 & 0.01 & 0.08 & 0.05 & 0.35 & 2.83 \\
\midrule
\multirow{7}{*}{{Open}} 
 & InternVL3.5-1B-HF & 0.05 & 0.02 & 0.02 & 0.27 & 0.05 & 0.17 & 0.15 & 0.56 & 2.44 \\
 & Qwen-3-VL-2B-Instruct & 0.07 & 0.02 & 0.02 & 0.31 & 0.04 & 0.18 & 0.15 & 0.59 & 2.36 \\
 & Qwen-3-VL-4B-Instruct & 0.01 & 0.01 & 0.01 & 0.20 & 0.01 & 0.11 & 0.12 & 0.56 & 2.93 \\
 & Qwen-3-VL-8B-Instruct & 0.01 & 0.01 & 0.01 & 0.16 & 0.01 & 0.10 & 0.09 & 0.53 & 2.51 \\
 & Google-Gemma-3-12B & 0.05 & 0.02 & 0.02 & 0.29 & 0.05 & 0.18 & 0.12 & 0.60 & 2.70 \\
 & Microsoft-Phi-4-Multimodal-Instruct & 0.06 & 0.02 & 0.02 & 0.30 & 0.06 & 0.18 & 0.12 & 0.60 & 2.82 \\
 & InternVL3.5-GPT-OSS-20B-A4B & 0.08 & 0.03 & 0.03 & 0.34 & 0.06 & 0.20 & 0.17 & 0.60 & 2.79 \\
 & Qwen-3-VL-30B-Instruct & 0.09 & 0.03 & 0.03 & 0.36 & 0.06 & 0.20 & 0.18 & 0.62 & 3.31 \\
 & Qwen-3-VL-235B-Instruct & 0.04 & 0.01 & 0.01 & 0.30 & 0.04 & 0.17 & 0.16 & 0.60 & \textbf{3.59} \\
\midrule
\multirow{3}{*}{{Thinking}} 
 & Qwen-3-VL-8B-Thinking & 0.02 & 0.01 & 0.01 & 0.25 & 0.03 & 0.14 & 0.13 & 0.58 & 2.81 \\
 & Qwen-3-VL-30B-Thinking & 0.03 & 0.01 & 0.01 & 0.26 & 0.03 & 0.15 & 0.15 & 0.59 & 3.21 \\
 & Qwen-3-VL-235B-Thinking & 0.03 & 0.01 & 0.01 & 0.27 & 0.04 & 0.16 & 0.15 & 0.60 & 3.43 \\
\midrule
\multirow{2}{*}{{DF}} 
 & SIDA-13B-Description & 0.01 & 0.01 & 0.01 & 0.24 & 0.03 & 0.16 & 0.15 & 0.58 & 2.32 \\
 & Qwen2.5-VL-Gen-Buster & 0.05 & 0.02 & 0.02 & 0.26 & 0.03 & 0.15 & 0.15 & 0.57 & 2.33 \\
\bottomrule
\end{tabular}

\label{tab:model_generation_metrics_revised}
\end{table*}

\begin{table*}[t]
\centering
\caption{Comparison of regression and correlation metrics across DeepfakeJudge-Meta and DeepfakeJudge-Meta-Human for pointwise evaluation. Arrows indicate desired direction of improvement (↑ higher is better, ↓ lower is better).}
\small
\begin{tabular}{l lccccccccc}
\toprule
\textbf{Type} & \textbf{Model} & \multicolumn{4}{c}{\textbf{DeepfakeJudge-Meta}} & \multicolumn{4}{c}{\textbf{DeepfakeJudge-Meta-Human}} \\
\cmidrule(lr){3-6}\cmidrule(lr){7-10}
 &  & RMSE\,($\downarrow$) & MSE\,($\downarrow$) & s\,($\uparrow$) & p\,($\uparrow$) & RMSE\,($\downarrow$) & MSE\,($\downarrow$) & s\,($\uparrow$) & p\,($\uparrow$) \\
\midrule
\multirow{2}{*}{Closed} 
 & Gemini-Flash-2.5 & 1.09 & 1.20 & 0.82 & 0.83 & 1.11 & 1.24 & 0.85 & 0.83 \\
 & GPT-4o-Mini & 0.78 & 0.60 & 0.87 & 0.87 & 0.81 & 0.66 & 0.88 & 0.86 \\
\midrule
\multirow{5}{*}{Open} 
 & Qwen-3-VL-2B-Instruct & 1.60 & 2.55 & 0.60 & 0.61 & 1.41 & 1.98 & 0.69 & 0.72 \\
 & Qwen-3-VL-4B-Instruct & 1.59 & 2.53 & 0.57 & 0.58 & 1.53 & 2.34 & 0.60 & 0.61 \\
 & Qwen-3-VL-8B-Instruct & 1.19 & 1.41 & 0.76 & 0.75 & 1.28 & 1.63 & 0.79 & 0.76 \\
 & Qwen-3-VL-30B-Instruct & 1.28 & 1.65 & 0.81 & 0.81 & 1.32 & 1.75 & 0.89 & 0.84 \\
 & Qwen-3-VL-235B-Instruct & 1.10 & 1.21 & 0.83 & 0.82 & 1.26 & 1.58 & 0.81 & 0.77 \\
\midrule
\multirow{3}{*}{Thinking} 
 & Qwen-3-VL-8B-Thinking & 1.09 & 1.19 & 0.83 & 0.82 & 1.18 & 1.39 & 0.84 & 0.81 \\
 & Qwen-3-VL-30B-Thinking & 1.02 & 1.05 & 0.83 & 0.83 & 1.04 & 1.07 & 0.85 & 0.83 \\
 & Qwen-3-VL-235B-Thinking & 1.01 & 1.02 & 0.84 & 0.84 & 0.95 & 0.91 & 0.85 & 0.86 \\
\midrule
\multirow{2}{*}{Ours} 
 & \textbf{DeepfakeJudge-3B} & \textbf{0.69} & \textbf{0.48} & \textbf{0.92} & \textbf{0.92} & \textbf{0.56} & \textbf{0.31} & \textbf{0.96} & \textbf{0.95} \\
 & \textbf{DeepfakeJudge-7B} & \textbf{0.61} & \textbf{0.37} & \textbf{0.94} & \textbf{0.93} & \textbf{0.50} & \textbf{0.25} & \textbf{0.96} & \textbf{0.95} \\
\bottomrule
\end{tabular}

\label{tab:dfj_metrics_revised}
\end{table*}

\section{Evaluation and Results}
We evaluate our approach along four axes. (1) We assess the reliability of VLMs for OOD-detection. (2) We evaluate the capability of vision-language models (VLMs) to generate accurate reasoning rationales, while also examining the reliability of current automatic metrics used to score reasoning quality. (3) We test the ability of the \textbf{DeepfakeJudge} \textbf{models (3B/7B)} to assess reasoning quality through pointwise and pairwise evaluation. (4) Finally, we measure the alignment between \textbf{DeepfakeJudge} and human judgments. Across all evaluations, we compare around fifteen state-of-the-art VLMs grouped into four model families: (1) closed-source VLMs (Gemini-2.5-Flash~\cite{comanici2025gemini25pushingfrontier}, GPT-4o-mini~\cite{openai_gpt4o_mini_2024}); (2) open-source VLMs (Qwen-3-VL~\cite{yang2025qwen3technicalreport}, InternVL~\cite{wang2025internvl35advancingopensourcemultimodal}, Gemma-3~\cite{gemmateam2025gemma3technicalreport}, Phi-4~\cite{abdin2024phi4technicalreport} variants); (3) reasoning-augmented open-source models (Qwen-3-VL-Thinking~\cite{yang2025qwen3technicalreport}); and (4) specialized deepfake detectors (SIDA~\cite{huang2025sidasocialmediaimage}, GenBuster++~\cite{wen2025busterxunifiedcrossmodalaigenerated}). Evaluation prompts are available in Appendix (Section \ref{tab:human_annotation_summary}).


\noindent\textbf{Deepfake Detection Evaluation.}
We begin by evaluating all models on the \textbf{DeepfakeJudge-Detect} benchmark to measure their capability to identify real versus fake images under out-of-distribution conditions. We report accuracy and F1-scores across real, fake, and overall categories.
From Table~\ref{tab:detect_revised}, closed-source models achieve strong performance on real images (Real F1 $\approx$ 70–74), but fail to generalize to fakes (Fake F1 $<$ 50). Open-source VLMs show moderate improvements, while reasoning-augmented models like Qwen-235B-Thinking reach better Fake F1 (79.8) yet remain inconsistent across domains. \textit{Qwen-3-VL-235B surpasses even State-of-the-Art closed source models, including Gemini-2.5-Flash and ChatGPT-4o-mini, by achieving an overall accuracy of 74.5\%.} Deepfake Detection models, such as SIDA and Gen-Buster++, fail to generalize to data generated by newer generation models.

\noindent\textbf{Reasoning Evaluation.}
We assess reasoning quality on the \textbf{DeepfakeJudge-Reason} dataset, where models generate textual explanations compared with human rationales using BLEU \cite{papineni-etal-2002-bleu} (n-gram precision), ROUGE \cite{lin-2004-rouge} (overlap-based recall), METEOR \cite{banerjee-lavie-2005-meteor} (semantic precision–recall F1), and BERTScore \cite{zhang2020bertscoreevaluatingtextgeneration} (embedding similarity). These metrics capture lexical fluency but fail to reflect visual grounding or factual accuracy. In contrast, \textbf{DeepfakeJudge-3B} evaluates explanations directly using the image, offering a more reliable signal of reasoning quality.

As shown in Table~\ref{tab:model_generation_metrics_revised}, all models achieve low BLEU and ROUGE scores (BLEU-3 $<$ 0.1), indicating weak alignment with human reasoning despite fluent text. Larger VLMs such as Gemini-2.5 and Qwen-3-VL-30B tend to produce verbose but generic justifications, while reasoning-augmented models offer only minor gains. In contrast, DeepfakeJudge scores correlate strongly with detection accuracy (Tables~\ref{tab:dfj_metrics_revised} and~\ref{tab:model_generation_metrics_revised}). Qwen-3-VL-235B attains the highest reasoning score (3.59), consistent with its top detection results, which shows that visual and semantic faithfulness, is key to reliable reasoning evaluation.

\noindent\textbf{Pointwise Evaluation.} We evaluate reasoning assessment quality using the \textbf{DeepfakeJudge-Meta} and \textbf{DeepfakeJudge-Meta-Human} datasets. We measure performance using root mean square error (RMSE) and mean square error (MSE) for regression accuracy, and Spearman ($s$) and Pearson ($p$) correlations to quantify alignment with human ratings. Lower RMSE and MSE, and higher correlation, indicate stronger agreement with humans. 
Table~\ref{tab:dfj_metrics_revised} shows that closed-source models perform reasonably (Gemini RMSE = 1.09, $p$ = 0.83), but open-source and reasoning-augmented models lag behind, often overestimating or underestimating reasoning quality. In contrast, DeepfakeJudge-3B and DeepfakeJudge-7B achieve substantial improvements. RMSE decreases to 0.61 and correlation rises to 0.94 on DeepfakeJudge-Meta, with even better results (RMSE = 0.50, $p = 0.95$) on DeepfakeJudge-Meta-Human, outperforming models that are more than \texttt{30X} their size.  These results show that our DeepfakeJudge's outputs are aligned with human's thinking rationale. 

\begin{table}[t]
\centering
\caption{Pairwise accuracy results for models on the DeepfakeJudge-Meta and DeepfakeJudge-Meta-Human datasets.}
\small
\begin{tabular}{l lcc}
\toprule
\textbf{Type} & \textbf{Model} & \textbf{DFJM} & \textbf{DFJMH} \\
\midrule
\multirow{2}{*}{Closed} 
 & Gemini-Flash-2.5 & 91.7 & 94.2 \\
 & GPT-4o-Mini & 90.3 & 89.8 \\
\midrule
\multirow{5}{*}{Open} 
 & Qwen-3-VL-2B-Instruct & 74.8 & 65.1 \\
 & Qwen-3-VL-4B-Instruct & 75.8 & 72.7 \\
 & Qwen-3-VL-8B-Instruct & 86.0 & 88.6 \\
 & Qwen-3-VL-30B-Instruct & 91.3 & 96.3 \\
 & Qwen-3-VL-235B-Instruct & 93.2 & 99.4 \\
\midrule
\multirow{3}{*}{Thinking} 
 & Qwen-3-VL-8B-Thinking & 89.2 & 93.2 \\
 & Qwen-3-VL-30B-Thinking & 92.5 & 97.7 \\
 & Qwen-3-VL-235B-Thinking & 90.8 & 95.5 \\
\midrule
\multirow{2}{*}{Ours} 
 & \textbf{DeepfakeJudge-3B} & \textbf{94.4} & \textbf{96.6} \\
 & \textbf{DeepfakeJudge-7B} & \textbf{96.2} & \textbf{98.9} \\
\bottomrule
\end{tabular}
\vspace{-2mm}
\vspace{-1mm}
\label{tab:dfj_accuracy_revised}
\end{table}

\noindent\textbf{Pairwise Evaluation.}
Finally, we assess relative reasoning preference through a \textbf{pairwise comparison} task, again on \textbf{DeepfakeJudge-Meta} and \textbf{DeepfakeJudge-Meta-Human}. This task emphasizes fine-grained discrimination of reasoning quality. Performance is measured by pairwise accuracy, which represents the percentage of model judgments that match human preferences.
As summarized in Table~\ref{tab:dfj_accuracy_revised}, DeepfakeJudge-7B achieves 96.2 percent pairwise accuracy on DeepFakeJudge-Meta and 98.9 percent on Deepfake-Judge-Meta-Human, surpassing both open- and closed-source models by large margins. Even much larger systems, such as Qwen-3-VL-235B (\texttt{30X} larger) and Chatgpt-4o-mini \cite{openai_gpt4o_mini_2024}, trail behind in human-consistent reasoning preference. 
These results confirm that our framework enables fine-grained, human-aligned judgment of reasoning fidelity.

\noindent\textbf{User Study.}
In our user study, we evaluated the quality of visual reasoning produced in our \textbf{DeepfakeJudge-Detect} dataset compared to \textbf{SIDA-Set-Description}\cite{huang2025sidasocialmediaimage}, \textbf{GPT-4o-mini}\cite{openai_gpt4o_mini_2024}, and \textbf{Qwen-3-VL-235B} \cite{yang2025qwen3technicalreport}, asking \textbf{ten} participants to choose which explanation best satisfied \textit{Faithfulness}, \textit{Groundedness}, \textit{Correctness}, \textit{Clarity}, and \textit{Usefulness}. Each annotator annotated 20 samples, which were sampled randomly from the overlapping set of SIDA-SET and DeefakeJudge-Meta dataset, and the samples were filtered for any NSFW content before annotation. 

Using majority voting, the DeepfakeJudge-Reason dataset's explanations were preferred in \(70\%\) of cases, while Qwen-3-VL-235B was chosen for 20\% images, and SIDA and Chatgpt-4o-mini \cite{openai_gpt4o_mini_2024} each for 5\% of the images. This shows that our pipeline-generated explanations were consistently seen as clearer, more accurate, and better grounded in the image content than those from other datasets. Importantly, rationales from the SIDA-dataset were the majority choice only once, indicating that its explanations were less aligned with user expectations, and were generated with over-reliance on text rather than the image itself. When comparing DeepfakeJudge-Meta and Qwen-3-VL-235B directly across 18 images, a binomial test (\(p \approx 0.015\)) confirms that DeepfakeJudge-Reason’s advantage is statistically significant, meaning users reliably preferred our explanations over Qwen’s reasoning rationales.



\vspace{-2mm}
\section{Conclusion}
\vspace{-2mm}

We present \textbf{DeepFakeJudge}, a unified framework for reasoning supervision and evaluation in deepfake detection.  
We construct an out-of-distribution benchmark covering generative and editing-based forgeries, a human-annotated reasoning dataset linking textual explanations to visual evidence, and a vision–language model trained as a reasoning judge.  
Through bootstrapped supervision, the framework scales human reasoning into structured ratings, enabling pointwise and pairwise evaluation of explanation quality.  
Comprehensive experiments show that our reasoning supervision achieves near-human correlation in reasoning assessment, thus paving the way for large-scale supervision of foundational models without the need for additional reasoning rationales. By combining human annotation, multimodal supervision, and automatic evaluation, this work establishes reasoning fidelity as a measurable dimension of trustworthy deepfake detection and provides a foundation for scalable, interpretable, verifiable, and generalizable supervision and evaluation of forensic systems.

\newpage
{
    \small
    \bibliographystyle{ieeenat_fullname}
    \bibliography{main}
}

\setcounter{page}{1}
\maketitlesupplementary

\begin{table*}[t]
\begin{center}
\scriptsize 
\begin{adjustbox}{max width=\linewidth}
\begin{tabular}{l|p{15cm}}
\multicolumn{2}{c}{\Large\textbf{Prompt filtering}} \\[2mm]
\multicolumn{2}{c}{\small\textbf{}} \\[2mm]
\toprule
\textbf{Category / Class} & \textbf{Associated Keywords} \\
\midrule
people-portrait &
portrait, headshot, close-up, candid, person, people, model, fashion, editorial, street portrait, wrinkles, freckles \\

people-activity &
athlete, runner, dancer, musician, chef, worker, doctor, engineer, teacher, construction, artisan, farmer, baker \\

nature-landscape &
mountain, valley, forest, desert, ocean, beach, river, waterfall, glacier, canyon, meadow, cliff, coast, island \\

animals-wildlife &
wildlife, bird, eagle, owl, lion, tiger, bear, wolf, fox, deer, elephant, giraffe, zebra, whale, dolphin \\

animals-pets &
dog, cat, puppy, kitten, hamster, parrot \\

architecture-exterior &
building, skyscraper, bridge, street, alley, facade, historic, monument, castle, church, mosque, temple, pagoda, synagogue, stadium \\

architecture-interior &
interior, living room, kitchen, bedroom, bathroom, office, workspace, studio, hotel lobby, museum, library, hallway \\

food-product &
dish, meal, cuisine, plate, bowl, garnish, restaurant, pastry, bread, cake, coffee, espresso, latte, burger, pizza, sushi, product photo, packaging, bottle, label \\

transportation &
car, train, bus, tram, bicycle, motorcycle, airplane, ship, harbor, terminal, station, highway, traffic, garage \\

sports-action &
soccer, football, basketball, tennis, boxing, martial arts, skateboard, surfing, skiing, snowboarding, cycling, swimming \\

macro-detail &
macro, close-up, texture, surface, dew, raindrops, insect, flower stamen \\

aerial-drone &
aerial, drone, overhead, bird's-eye view, top-down \\

night-lowlight &
night, low light, neon, long exposure, light trails, starry sky, milky way \\

weather &
rain, snow, fog, storm, lightning, hail, mist, clouds \\

underwater &
underwater, coral, reef, scuba, snorkel, sea turtle \\

events &
wedding, festival, concert, parade, market, ceremony \\

fashion-beauty &
runway, couture, makeup, hairstyle, wardrobe, studio portrait \\

documentary-street &
street photography, documentary, photojournalism, everyday life, candid \\

\midrule
\textbf{unreal (negative)} &
dragon, fairy, wizard, elf, demon, vampire, alien, robot, cyberpunk, anime, cartoon, 3d render, digital art, surreal, concept art, illustration, ai-generated, futuristic, mythical, glowing, hologram, dreamscape \\

\textbf{nsfw (negative)} &
nsfw, nude, explicit, erotic, fetish, sexual, breasts, lingerie, bondage, underage, gore, blood, decapitated, mutilated, NSFL \\
\bottomrule
\end{tabular}
\end{adjustbox}
\end{center}
\caption{ Category and negative-class keyword sets used for prompt filtering. 
We evaluate the prompts from \citet{wangDiffusionDBLargescalePrompt2022} using a weighted scoring function that emphasizes linguistic realism and photographic relevance. The final score combines three main components: prompt length (60\% weight, modeled with a sigmoid favoring 30 to 100 words), clause count (30\% weight), and a +0.5 bonus if any term from the photographic keyword whitelist appears. Additional penalties were applied for overly long or repetitive text. The resulting scores ranged from 0.70 to 1.38 (median $\approx$~1.19), with higher values common since most prompts triggered the photo bonus. From 2{,}000 exported samples (columns: \texttt{prompt}, \texttt{category}, \texttt{score}), the average prompt length was about 55~words ($\sigma \approx 9$; range 45 to 122), and 84.9\% of prompts contained explicit photographic terms. Thirteen categories were represented: \texttt{people-portrait} dominated with 1{,}697 prompts (84.9\%), followed by \texttt{nature-landscape} (77), \texttt{transportation} (72), and smaller groups such as \texttt{animals-pets} (58) and \texttt{events} (15). Mean category scores varied, with \texttt{people-portrait} averaging~1.218 and others clustering around~0.71 to 0.77, reflecting the bias toward photographic realism.}
\label{tab:prompt-filtering}
\end{table*}

\clearpage

\begin{table*}[htbp]
\footnotesize
\begin{center}
\begin{adjustbox}{max width=\linewidth}
\begin{tabular}{p{0.15\linewidth} p{0.3\linewidth} p{0.50\linewidth}}
\multicolumn{3}{c}{\Large\textbf{Human Annotation}} \\[2mm]

\toprule
\textbf{Flag Name} & \textbf{Check} & \textbf{Pass / Fail Description} \\
\midrule

Shadows & 
Do all shadows point roughly the same way, show a dark contact at touchpoints, and get softer as they stretch away? &
\textbf{PASS:} All outdoor shadows lean the same way; tight dark line under shoes; edges soften outward. 
\textbf{FAIL:} One shadow points the opposite way; feet ``float'' (no dark contact); shadow edge equally sharp everywhere. \\

Lighting Match &
Are objects lit from the same side with similar brightness/contrast for the scene? &
\textbf{PASS:} Faces, walls, and props all brighter on the left side. 
\textbf{FAIL:} One object bright from the left while nearby objects are bright from the right. \\

Color Cast Consistency &
Do whites/neutrals share the same warm/cool tint? &
\textbf{PASS:} All whites slightly warm indoors. 
\textbf{FAIL:} A shirt is cold blue while everything else looks yellowish. \\

Relative Size / Scale &
Do familiar things get smaller with distance and look believably sized? &
\textbf{PASS:} Far people/cars noticeably smaller than near ones. 
\textbf{FAIL:} A far person is as big as someone close. \\

Perspective / Straight-Line Convergence &
Do parallel lines (road edges, table sides) aim toward the same point(s)? &
\textbf{PASS:} Building edges converge consistently. 
\textbf{FAIL:} An added sign's edges converge to a different point. \\

Front--Back Overlap &
Does the front object cleanly cover the back object without halos or see-through? &
\textbf{PASS:} Vase cleanly blocks the book behind it. 
\textbf{FAIL:} Background leaks through hair/edges; color fringe halo. \\

Contact \& Support &
Do objects look supported and press into soft things when they should? &
\textbf{PASS:} Cushion dips under a person; grass bends under a foot. 
\textbf{FAIL:} Heavy box sits on a blanket with zero dent. \\

Focus / Blur with Depth &
Does blur increase smoothly with distance from the focus plane? &
\textbf{PASS:} Foreground soft $\rightarrow$ subject sharp $\rightarrow$ background gently soft. 
\textbf{FAIL:} A pasted object stays razor-sharp while neighbors at the same depth are blurry. \\

Reflections \& Transparency &
Do reflections show the right thing at the right angle, and do transparent surfaces show both some reflection and some see-through? &
\textbf{PASS:} Person appears (darker) in a shop window at the expected position; window shows faint outdoor reflection plus the room behind. 
\textbf{FAIL:} No reflection where expected, wrong pose/angle in mirror, or window behaves as a perfect mirror or perfectly clear at a glancing angle. \\

Material Shine &
Do shiny things have crisp highlights and matte things have soft, broad highlights? &
\textbf{PASS:} Metal = bright, sharp highlight; fabric = soft highlight. 
\textbf{FAIL:} Everything has the same plasticky shine. \\

Texture / Pattern Follow-Through &
Do stripes, weaves, or wood grain bend and scale with folds and curves? &
\textbf{PASS:} Shirt stripes curve smoothly around the sleeve. 
\textbf{FAIL:} Pattern stays flat or smeared across bends. \\

Text \& Small Details &
Are letters/numbers legible and small parts coherent? &
\textbf{PASS:} Street sign letters readable, normal spacing. 
\textbf{FAIL:} Gibberish letters, melted numbers, warped watch face markings. \\

Object Completeness \& Counts &
Are obvious parts present and counted right (fingers, spokes, chair legs, petals, etc.)? &
\textbf{PASS:} Hand shows five distinct fingers; bicycle has two wheels and believable spokes. 
\textbf{FAIL:} Wrong counts (e.g., three fingers), missing chair leg, duplicated petal, merged limbs. \\

Edges \& Boundaries (cut-out / halo check) &
Do object edges look like they naturally belong in the scene? Edges should be as sharp/soft as nearby stuff at the same distance, with no glowing outline, no weird color rim, and no ``scissor-cut'' look. &
\textbf{PASS:} Jacket sleeve edge slightly soft like nearby edges; wispy hair/fur irregular (not clumped); no bright outline; edge grain matches background. 
\textbf{FAIL:} Bright/dark halo; too-straight cut line; color fringe; jaggy edge; \\

\bottomrule
\end{tabular}
\end{adjustbox}
\end{center}
\caption{Taxonomy of the flags used for human annotation}
\label{tab:flag-taxonomy}
\end{table*}

\clearpage

\label{sec: qualitative_examples}

\begin{table*}[h]
\begin{center}
    \begin{adjustbox}{max width=\linewidth}
    \begin{tabular}{l|p{15cm}}
    
    \multicolumn{2}{c}{\LARGE \textbf{Qualitative Examples for degraded reasoning responses.}} \\[2mm]
    \multicolumn{2}{c}{\LARGE \textbf{}} \\[2mm]

    \multicolumn{2}{c}{\includegraphics[width=0.4\linewidth]{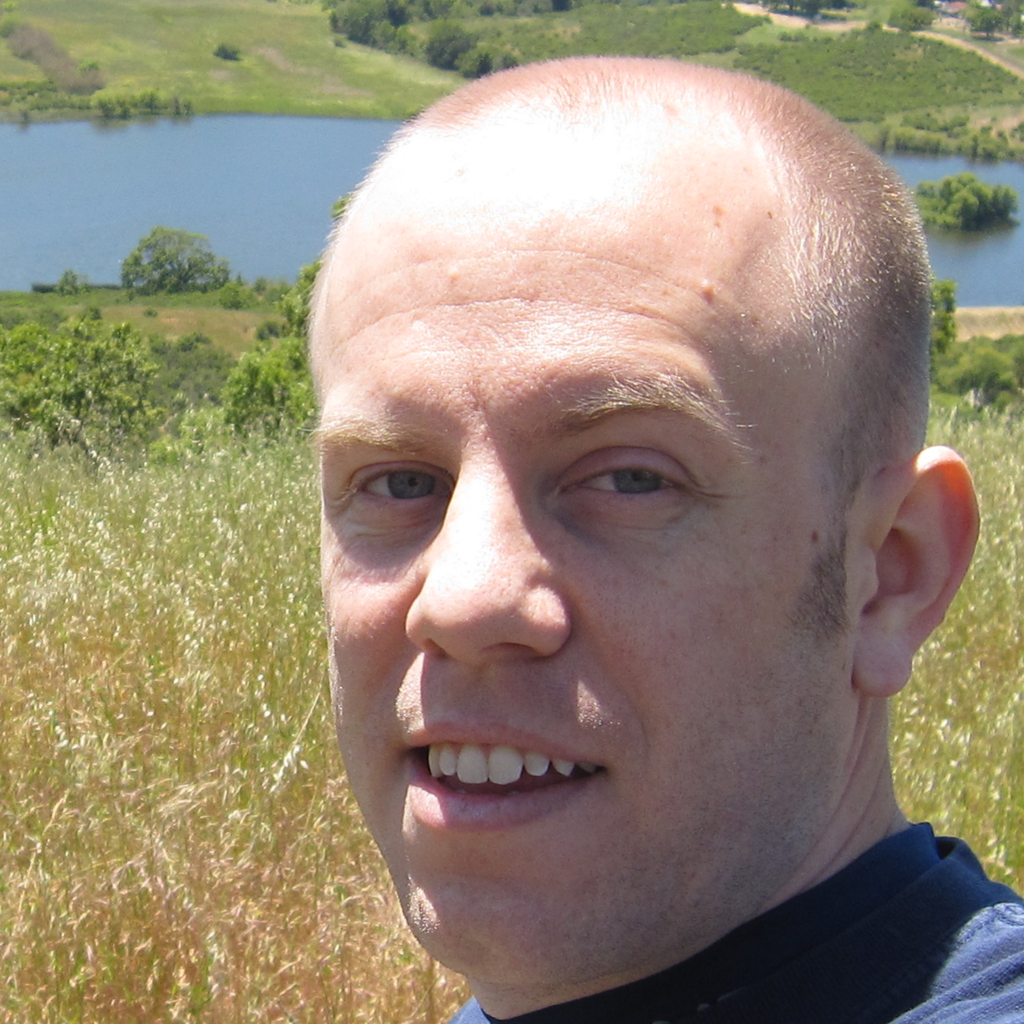}} \\
    \multicolumn{2}{c}{\small (Image label: Real)} \\
    \toprule
    \textbf{Rating} & \textbf{Reasoning} \\
    \midrule
    5 (Gold) & The bright sunlight casts a sharp shadow on the man’s face, with clear shading under his nose and around his eyes. The scalp’s highlights and soft skin texture show natural variation and pores. Background grass and water blur smoothly with depth, contrasting the sharp facial features in the foreground. The size and perspective of the man’s head relative to the background landscape match a close-up view with distant elements. \\
    4 & The sunlight produces distinct shadows on the man's face, especially under his nose and near his eyes, which aligns with typical outdoor lighting. The scalp displays natural highlights, while the skin texture shows pores and slight variations. The background, including grass and water, has a smooth blur that matches the depth of field, complementing the sharpness of the face in the foreground. The head's size and perspective are consistent with the distant landscape, supporting the notion that this is an authentic close-up shot. \\
    3 & The image exhibits natural lighting, with shadows on the face and highlights on the scalp, although certain skin details seem a bit soft or slightly overexposed. The background blurs gently, consistent with a shallow depth of field, and the head size appears reasonable compared to the landscape. There are no apparent artifacts or edges that indicate manipulation, but the image lacks finer details in some regions. \\
    2 & While the face shows shadows and highlights, some skin areas appear unusually smooth or flat, hinting at potential retouching. The background blur is noticeable but inconsistent, with certain grass areas sharper than one would expect. Furthermore, the scale of the head seems slightly disproportionate to the distant scenery, and there are subtle blending artifacts around the ears. These details suggest that the image may not be entirely authentic. \\
    1 & The lighting on the face is not uniform, with shadows that do not correspond to the sunlight direction in the background. The scalp has an odd shine that resembles plastic, and the skin texture is devoid of natural pores. The background blur is patchy and inconsistent, and the head seems to be placed onto the landscape with unclear edges. These signs strongly indicate that the image is either fake or extensively edited. \\
    \bottomrule
    \end{tabular}
    \end{adjustbox}
\end{center}
\caption{Example of a real image and corresponding degraded ratings produced by our bootstrapping process.}
\end{table*}

\begin{table*}[t]
\begin{center}
    \begin{adjustbox}{max width=\linewidth}
    \begin{tabular}{l|p{15cm}}
    \multicolumn{2}{c}{\includegraphics[width=0.4\linewidth]{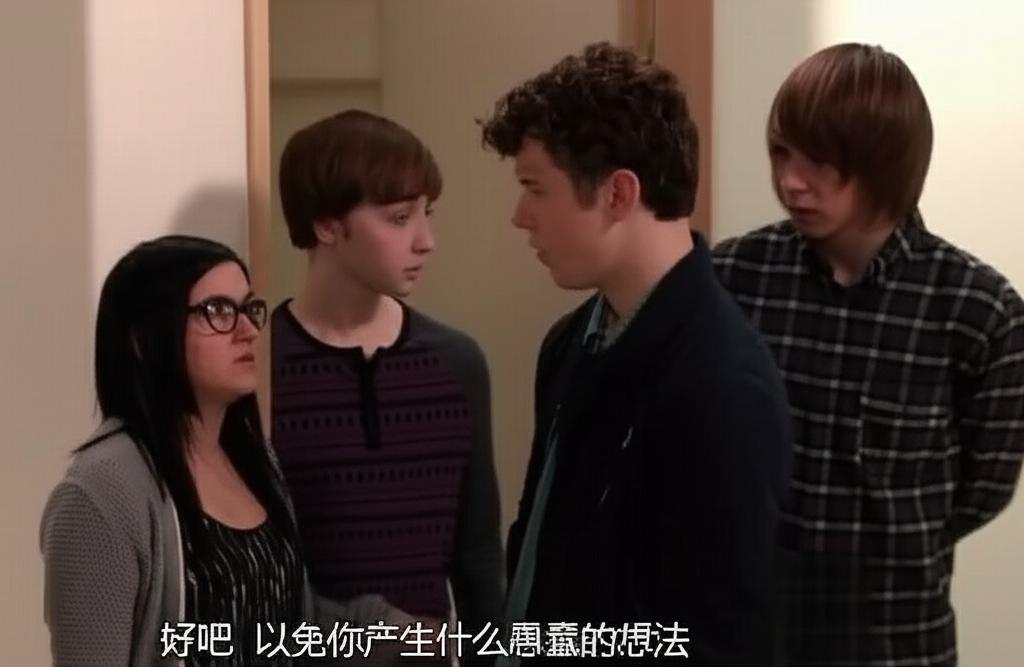}} \\
    \multicolumn{2}{c}{\small (Image Label: Edited)} \\
    \toprule
    \textbf{Rating} & \textbf{Reasoning} \\
    \midrule
    5 (Gold) & The man in the center maintains a direct and intense gaze rather than lowering his eyes, which is unexpected given the context that suggests remorse or conflict. His expression lacks subtle emotional cues, such as softness or vulnerability, that might indicate regret. The overall posture and facial alignment seem slightly stiff, detracting from a natural, relatable interaction. The other individuals’ expressions and body language appear more neutral, creating a contrast that highlights the central figure’s unusual direct stare. This inconsistency in expression and gaze contributes to a sense of artificiality, suggesting the image may have been generated or manipulated. \\
    4 & The central figure's gaze is remarkably direct and confrontational, which is inconsistent with what one might anticipate from someone who is remorseful or anxious. His facial expression seems rigid and devoid of the emotional nuance found in authentic remorse. Meanwhile, the surrounding individuals exhibit more neutral and subdued expressions, which accentuates the central figure's unusual stare. This incongruity in emotional expressions and gaze suggests that the image may have been manipulated. \\
    3 & The man in the center looks directly at the camera with a strong gaze, which feels a bit unusual for the scene context. His face seems a little stiff compared to the others, who appear more relaxed or neutral. This difference makes the image feel less natural, though it's not fully clear if it's edited or not. \\
    2 & The eyes of the primary figure appear oddly intense, and the expression seems unnatural, while the others look fine. The direct gaze could suggest editing, but there are no clear signs of manipulation in other areas. It's possible this is an authentic image with just a peculiar expression or lighting that makes it look strange. \\
    1 & The man is looking straight at the camera and smiling, which is perfectly normal. There is no evidence of manipulation or synthetic content, and the expressions of the other people are in harmony with the environment. All elements seem authentic, with uniform lighting and natural stances. \\
    \bottomrule
    \end{tabular}
    \end{adjustbox}
\end{center}
\caption{Example of an edited image and corresponding degraded ratings produced by our bootstrapping process.}
\end{table*}

\begin{table*}[t]
\begin{center}
    \begin{adjustbox}{max width=\linewidth}
    \begin{tabular}{l|p{15cm}}
    \multicolumn{2}{c}{\includegraphics[width=0.4\linewidth]{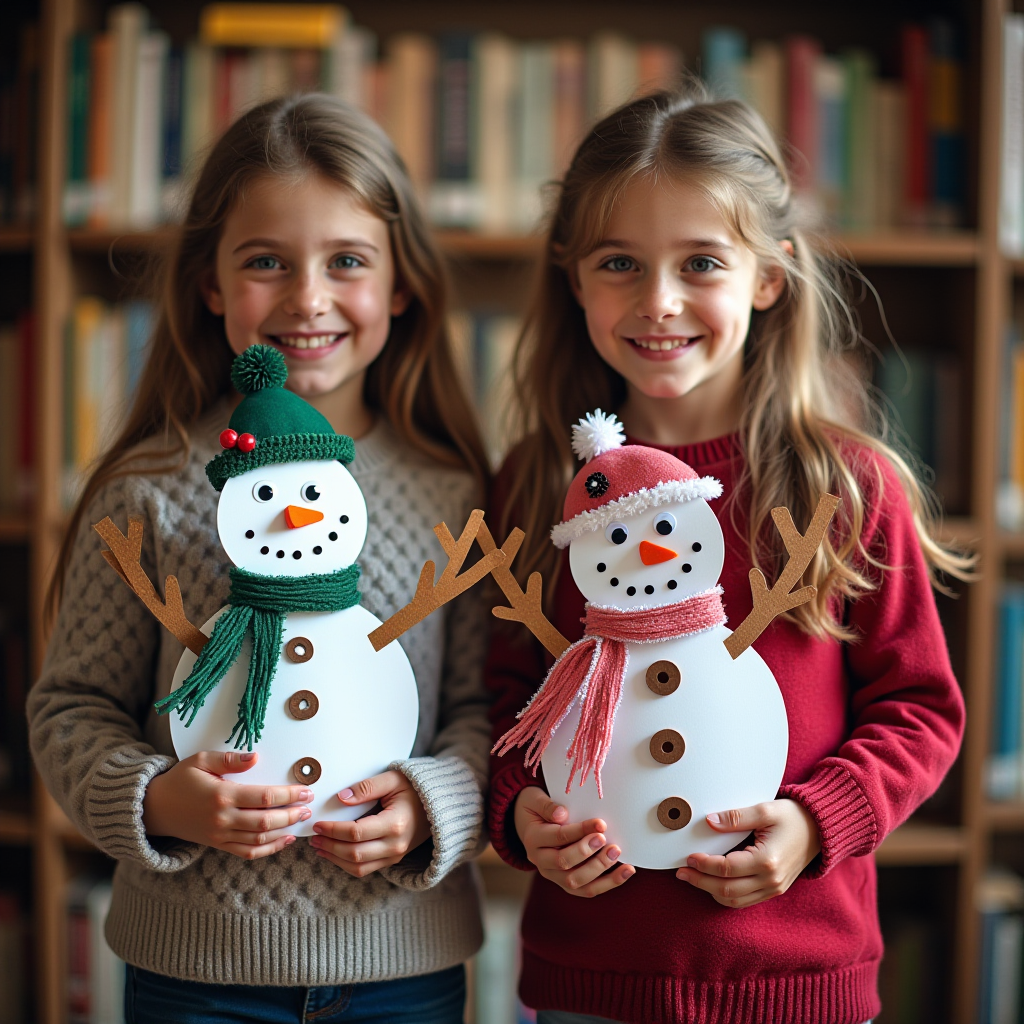}} \\
    \multicolumn{2}{c}{\small (Image Label: Fake)} \\
    \toprule
    \textbf{Rating} & \textbf{Reasoning} \\
    \midrule
    5 (Gold) & The skin on both girls looks overly smooth and lacks natural pores or texture, appearing unnaturally bright and plastic-like across their faces and hands. One of the fingers, particularly the left girl's hand in the lower center, is formed irregularly with an unnatural shape, disrupting the normal anatomy of the hand. The fingers seem merged or misshapen, breaking the expected geometry. Lighting on the skin is consistently even, which contrasts with the subtle shadows usually found in photographs. Additionally, the boundaries of the fingers and hands look somewhat blurred and unnatural compared to the crisp edges of the background elements. These flaws indicate the image is AI-generated. \\
    4 & The skin texture on the girls looks unnaturally smooth and bright, devoid of typical facial pores and natural skin variations. The left girl's hand, especially near the fingers, shows irregular shapes and merging that do not align with normal anatomy. The lighting is very uniform, lacking subtle natural shadows, and the edges around the fingers appear slightly blurred in contrast to the sharper background. These minor inconsistencies suggest that the image is probably AI-generated or heavily manipulated. \\
    3 & This image features both girls with skin that is unnaturally smooth, which looks somewhat unrealistic. Additionally, one finger on the left girl appears to be malformed or merged, not quite resembling a typical hand. The lighting is consistent and a bit flat, while the edges of the fingers are not very defined. These characteristics suggest that the image may have been digitally produced or manipulated, although some details remain unclear. \\
    2 & While the girls look largely authentic, certain parts of their skin seem overly smooth and bright, which might be due to lighting or editing. The left girl's fingers do not appear properly formed, possibly due to an awkward angle or blur. The overall lighting remains consistent, but the skin gives off a somewhat plastic-like impression. I am unsure if this image is genuine or artificial, but it appears to be edited. \\
    1 & The girls are happily holding snowmen and smiling. The lighting seems natural, and the bookshelves in the background are clear. The fingers and skin appear normal, lacking any obvious flaws or blurriness. This image resembles a standard photograph without evidence of manipulation or AI generation. \\
    \bottomrule
    \end{tabular}
    \end{adjustbox}
\end{center}
\caption{Example of a fake image and corresponding degraded ratings produced by our bootstrapping process.}
\end{table*}

\begin{table*}[t]
\begin{center}
    \begin{adjustbox}{max width=\linewidth}
    \begin{tabular}{l|p{15cm}}
    \multicolumn{2}{c}{\includegraphics[width=0.4\linewidth]{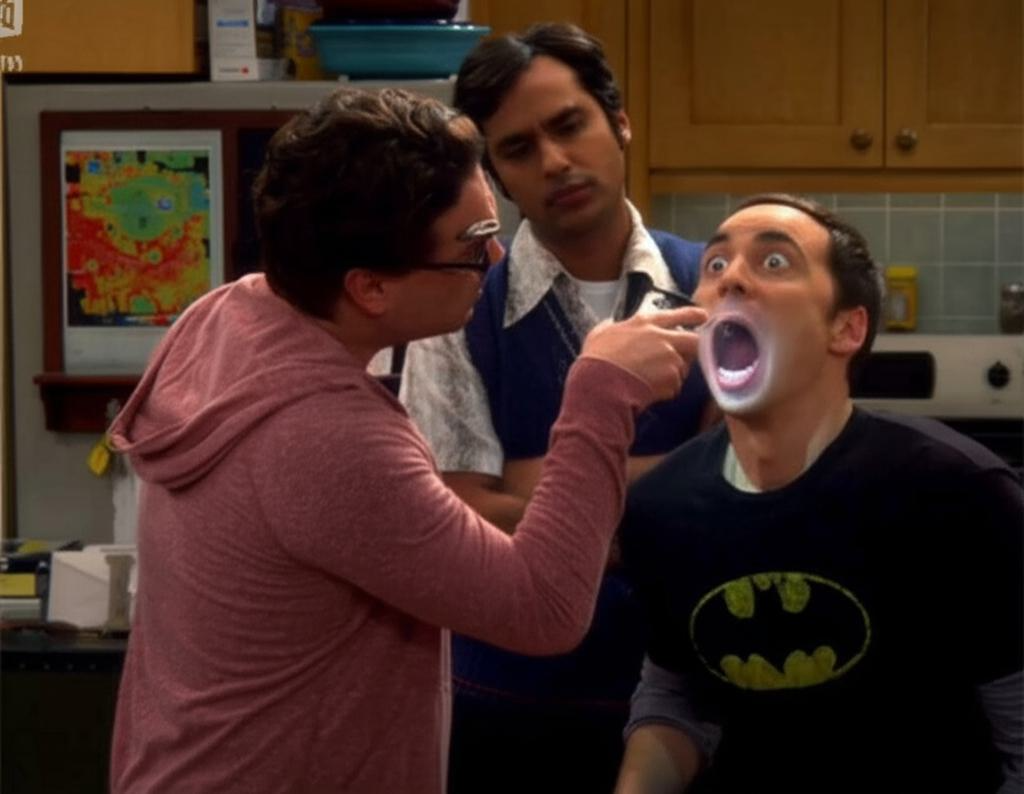}} \\
    \multicolumn{2}{c}{\small (Image Label: Edited)} \\
    \toprule
    \textbf{Rating} & \textbf{Reasoning} \\
    \midrule
    5 (Gold) & The eyes of the man on the right appear unnaturally warped and lack realistic detail, contributing to a disturbing facial expression. The intersection between the spectacle and the face of the man on the left shows unnatural blending, with parts of the glasses and skin merging in an unrealistic way. This overlap issue creates an impression of transparency rather than proper layering. The lighting and edges around the glasses and facial features also look inconsistent, causing a blurred boundary effect. These distorted eyes and implausible overlaps are clear signs that the image was generated by AI. \\
    4 & The individual on the right has eyes that are distinctly distorted with unnatural shapes, and his mouth is overly open, which looks unrealistic. Moreover, the spectacles on the left man's face do not blend effectively; the edges around the glasses seem to leak into the skin, causing an unusual transparency effect. These overlapping artifacts and the irregularity of the eyes suggest that the image is either AI-generated or has been heavily manipulated. \\
    3 & The person on the right has eyes that look strange and unrealistic, and the glasses on the left seem to blend improperly with the face. The lighting around the glasses and the edges of the face is uneven, which makes the image seem somewhat unnatural. These elements suggest possible manipulation or synthetic modification, even if the exact problems are not distinctly outlined. \\
    2 & The eyes on the right appear unusual, possibly due to the expression or lighting conditions. The glasses on the left individual seem somewhat unclear where they connect to the face, which might indicate a quality issue with the photograph. While there are signs of potential editing, the overall scene looks relatively normal. \\
    1 & The image appears generally normal; the eyes and glasses look appropriate without noticeable distortions or blending issues. The lighting is consistent across the faces and objects in the scene. It resembles a casual photograph with no indications of manipulation or artificiality. \\
    \bottomrule
    \end{tabular}
    \end{adjustbox}
\end{center}
\caption{Example of an edited image and corresponding degraded ratings produced by our bootstrapping process.}
\end{table*}

\begin{table*}[t]
\begin{center}
    \begin{adjustbox}{max width=\linewidth}
    \begin{tabular}{l|p{15cm}}
    \multicolumn{2}{c}{\includegraphics[width=0.4\linewidth]{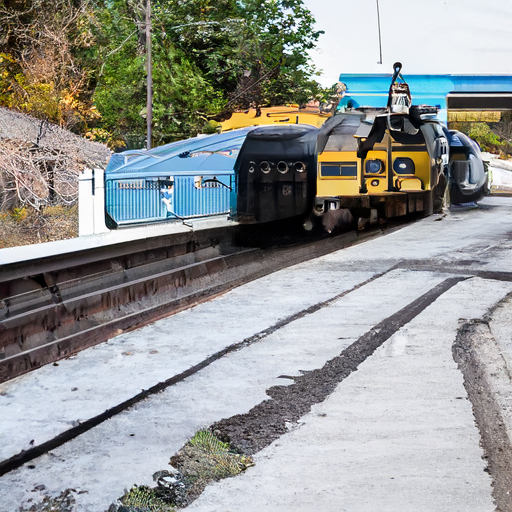}} \\
    \multicolumn{2}{c}{\small (Image Label: Fake)} \\
    \toprule
    \textbf{Rating} & \textbf{Reasoning} \\
    \midrule
    5 (Gold) & The train in the mid-right appears unnaturally bright and overly smooth, lacking realistic surface textures that would normally show wear and detail. The entire upper-right train section is distorted, with shapes fused together and unclear details, making it hard to distinguish individual parts. The merging of elements creates an unnatural and confusing structure that does not align with typical train forms. Edges between the train and the background are uneven and lack depth, further weakening the image's realism. These texture and structural anomalies clearly indicate that the scene is artificially generated. \\
    4 & The train located on the right side is strikingly bright and shows a lack of detailed texture, giving it an overly smooth and synthetic appearance. In the upper-right section, the shapes seem to merge, complicating the identification of distinct train parts. The boundaries between the train and its background are somewhat uneven, further diminishing its natural look. These texture irregularities and the blending of forms strongly suggest that the image is AI-generated. \\
    3 & The train exhibits an unusually bright and smooth texture, with certain sections blending into one another in the upper right area. The distinct features of the train's structure are somewhat obscured, and the edges around it lack sharpness, indicating possible alterations. Overall, the image raises suspicion and does not appear entirely authentic, despite some visible details. \\
    2 & The train looks a little off because some parts seem merged and unclear, especially on the upper right. The colors also look a bit unnatural. However, the rest of the scene looks okay. It might be edited or just a low-quality photo. There are no strong clear signs, so I'm not completely sure. \\
    1 & The train looks normal and the tracks appear fine, with no visible issues. The colors and textures seem consistent with a real photo. The background and surrounding objects also look natural. Nothing stands out as fake or edited here, so this image is definitely real. \\
    \bottomrule
    \end{tabular}
    \end{adjustbox}
\end{center}
\caption{Example of fake image and corresponding degraded ratings produced by our bootstrapping process..}
\end{table*}

\begin{table*}[t]
\begin{center}
    \begin{adjustbox}{max width=\linewidth}
    \begin{tabular}{l|p{15cm}}
    \multicolumn{2}{c}{\includegraphics[width=0.4\linewidth]{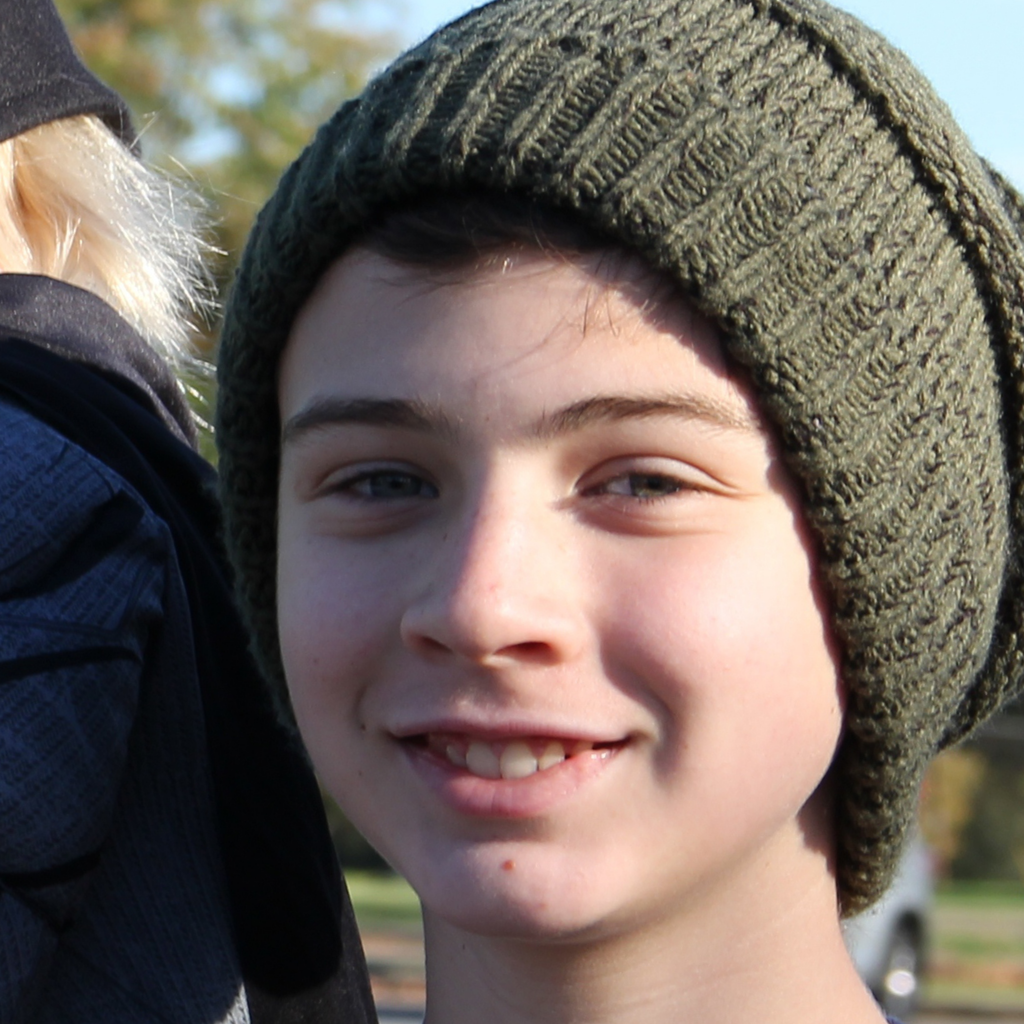}} \\
    \multicolumn{2}{c}{\small (Image label: Real)} \\
    \toprule
    \textbf{Rating} & \textbf{Reasoning} \\
    \midrule
    5 (Gold) & Soft, natural daylight uniformly illuminates the face and the knit hat, creating subtle shadows beneath the eyebrows and nose that correspond with the light source. The knitting texture on the hat is consistent in scale and shows realistic stitches and shadows. The face's sharpness contrasts smoothly with the slight blur of the background elements, providing depth. The hair near the hat edges blends naturally without halos, and the skin tones show natural variation in hue and lighting. \\
    4 & The lighting is gentle and natural, providing even daylight across both the face and the knit hat. Shadows beneath the eyebrows and nose align appropriately with the light source's direction. The knitting texture on the hat displays clear, realistic stitch details and a consistent scale. The face is sharply focused, contrasting with the slightly blurred background, which adds depth. The hair edges adjacent to the hat blend naturally without visible halos. Skin tones show credible variations in color and lighting, supporting the image's authenticity. \\
    3 & The image shows a daylight scenario that generally looks natural, with mild shadows present on the face and hat. The knit texture of the hat is apparent, though some areas lack detail. The face is sharply focused while the background is somewhat blurred, enhancing the depth. Hair along the edges of the hat blends reasonably well, although a few spots could be smoother. Skin tones appear normal, but subtle inconsistencies might stem from lighting or minor edits. Overall, the image seems real, though not completely flawless. \\
    2 & The illumination on the face and hat is uneven, with certain shadows looking inconsistent or too sharp for natural daylight. The knit texture on the hat is visible but appears somewhat flat in certain areas, suggesting possible retouching. The background blur is irregular, and some hair edges near the hat seem slightly unnatural or haloed. Skin tones vary, but some regions look excessively smoothed or edited. These factors cast doubt on the image's complete authenticity, though it may not be heavily altered. \\
    1 & The lighting is mismatched, with shadows on the face going in different directions and inconsistent with the outdoor background. The knit hat texture looks artificial and overly uniform, lacking natural stitch variation. Hair edges around the hat show clear halos and blending errors. The background blur is uneven, and skin tones appear overly airbrushed or colored unnaturally. These problems strongly indicate the image has been digitally manipulated or is fake. \\
    \bottomrule
    \end{tabular}
    \end{adjustbox}
\end{center}
\caption{Example of a real image and corresponding degraded ratings produced by our bootstrapping process..}
\end{table*}

\clearpage

\label{section: paraphrase_qualitative}


\begin{table*}[t]
\centering
\renewcommand{\arraystretch}{1.15}
\setlength{\tabcolsep}{6pt}
\begin{tabularx}{\linewidth}{@{}S{0.05}S{0.95}@{}}
\multicolumn{2}{c}{\LARGE \textbf{Qualitative Examples of Paraphrased Gold-Standard Ratings.}} \\[2mm]
\multicolumn{2}{c}{\LARGE \textbf{}} \\[2mm]
\multicolumn{2}{c}{\includegraphics[width=0.4\linewidth]{images/example_images/para_ex2.png}} \\
\multicolumn{2}{c}{\small (Image Label: Edited)} \\
\toprule
\textbf{Variant} & \textbf{Reasoning} \\
\midrule
Gold & The man’s face and head area exhibit unnatural texture, appearing synthetic and inconsistent with typical skin and hair details. Additionally, the motorcycle seat shows a dip as if supporting weight, yet the man is seated further forward, not making contact with that indentation. These discrepancies in texture and physical contact indicate manipulation in these areas. The rest of the image, including the motorcycle, background, and rider’s clothing, appears visually consistent and realistic. \\
Para-1 & The man's face texture looks somewhat unnatural, with an inconsistent pattern that suggests synthetic alteration. Also, the motorcycle seat shows a clear dip as if bearing weight, but the man is seated slightly forward, not pressing down on that area. These visual clues suggest the face region was edited. The rest of the scene, including the motorcycle and background, seems consistent and natural. \\
Para-2 & The texture of the man's face appears somewhat artificial, exhibiting an inconsistent pattern that indicates possible synthetic modification. Additionally, the motorcycle seat shows a noticeable dip as if someone is applying pressure, yet the man is positioned slightly forward, not affecting that area. These visual indicators imply that the face region has been altered. The remainder of the scene, including the motorcycle and background, appears coherent and natural. \\
Para-3 & The texture on the man's face seems oddly synthetic, with a pattern that lacks consistency, hinting at digital alteration. Furthermore, the motorcycle seat exhibits a distinct depression, suggesting weight is applied, while the man is seated slightly forward, not influencing that section. These visual signs suggest that the facial area has been modified. The rest of the scene, such as the motorcycle and background, looks consistent and realistic. \\
Para-4 & The man's facial texture appears somewhat unnatural, with a pattern that suggests it may have been synthetically altered. Moreover, the motorcycle seat displays a clear indentation, implying that weight is being exerted, yet the man is leaning slightly forward, not applying pressure to that area. These visual cues indicate that the face region has undergone editing. The rest of the scene, including the motorcycle and background, seems consistent and authentic. \\
\bottomrule
\end{tabularx}
\caption{Comparison of a fake image and the corresponding paraphrased and Gold-standard reasoning produced by our bootstrapping process.}
\end{table*}

\begin{table*}[t]
\centering
\renewcommand{\arraystretch}{1.15}

\setlength{\tabcolsep}{6pt}
\begin{tabularx}{\linewidth}{@{}S{0.05}S{0.95}@{}}
\multicolumn{2}{c}{\includegraphics[width=0.4\linewidth]{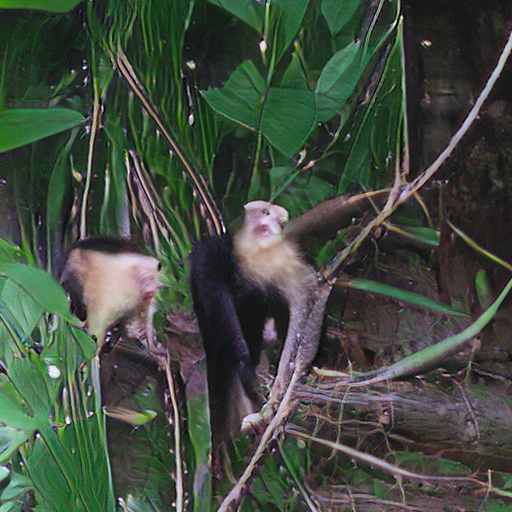}} \\
\multicolumn{2}{c}{\small (Image Label: Fake)} \\
\toprule
\textbf{Variant} & \textbf{Reasoning} \\
\midrule
Gold & Two animals blend confusingly into the surrounding foliage, lacking clear form or recognition, with edges smudged against leaves and branches. The front leaves exhibit inconsistent lighting compared to the overall scene, appearing artificially bright with an uneven glow that doesn't match the ambient light direction. Shadows under the animals are absent, resulting in unnatural floating effect and weak grounding. The perspective appears compressed in the lower-right, making the background look too close and flattened, disrupting spatial depth cues. These issues with object blending, lighting disparity, missing shadows, and distorted depth strongly indicate the image is AI-generated. \\
Para-1 & The image shows two animals poorly defined and blending into the dense foliage, with unclear boundaries that reduce recognizability. The lighting on the leaves in the foreground is uneven and doesn't align well with the rest of the scene, suggesting inconsistent illumination. Notably, shadows beneath the animals are missing, causing them to appear as if floating above the ground. The perspective in the lower-right section looks unnaturally compressed, making the background seem too near and flattening the spatial depth. These subtle inconsistencies in object clarity, lighting, shadows, and perspective point toward an artificially generated image. \\
Para-2 & The image depicts two animals that are not clearly defined and are blending into the surrounding foliage, resulting in indistinct edges that hinder identification. The lighting on the foreground leaves is inconsistent and does not match the overall scene, indicating uneven illumination. Additionally, the shadows beneath the animals are absent, making them appear as if they are floating. The perspective in the lower-right area appears unnaturally compressed, causing the background to seem too close and flattening the depth of the scene. These subtle discrepancies in object definition, lighting, shadows, and perspective suggest that the image is artificially generated. \\
Para-3 & This image features two animals that are poorly defined and merge with the dense foliage, leading to unclear boundaries that make recognition difficult. The lighting on the leaves in the foreground is uneven and does not correspond well with the rest of the scene, suggesting inconsistency in illumination. Notably, the absence of shadows under the animals creates an illusion of them floating above the ground. The perspective in the lower-right section appears unnaturally squished, making the background look too close and diminishing spatial depth. These minor inconsistencies in object clarity, lighting, shadows, and perspective indicate that the image is likely artificially generated. \\
Para-4 & The image shows two animals that are not distinctly visible and blend into the thick foliage, resulting in unclear outlines that impede recognition. The lighting on the leaves in the foreground is inconsistent and does not align with the overall scene, hinting at uneven illumination. Importantly, the shadows beneath the animals are missing, giving them a floating appearance. The perspective in the lower-right corner seems unnaturally compressed, causing the background to appear too close and flattening the spatial depth. These subtle inconsistencies in object definition, lighting, shadows, and perspective suggest that this image is artificially generated. \\
\bottomrule
\end{tabularx}
\caption{Comparison of a fake image and the corresponding paraphrased and Gold-standard reasoning produced by our bootstrapping process.}
\end{table*}

\begin{table*}[t]
\centering
\renewcommand{\arraystretch}{1.15}
\setlength{\tabcolsep}{6pt}
\begin{tabularx}{\linewidth}{@{}S{0.05}S{0.95}@{}}
\multicolumn{2}{c}{\includegraphics[width=0.4\linewidth]{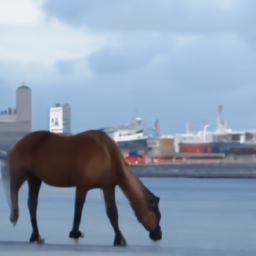}} \\
\multicolumn{2}{c}{\small (Image Label: Fake)} \\
\toprule
\textbf{Variant} & \textbf{Reasoning} \\
\midrule
Gold & The horse appears to be missing its left hind leg, resulting in a strange, incomplete posture. Its ears and face display inconsistent shapes and edges that do not align naturally, lending a distorted appearance. Despite the reflective surface of the water beneath, the horse's reflection is entirely absent, which breaks expected visual continuity. Additionally, the horse seems to be walking directly on the water's surface rather than on solid ground, creating an unrealistic effect of suspension. These geometric inconsistencies, lack of proper reflections, and object incompleteness strongly suggest that the image was AI-generated. \\
Para-1 & The horse's left hind leg is notably absent, causing an unnatural pose. The ears and face show some irregularities in shape that suggest digital alteration. The reflection on the water is missing where it should appear, breaking the expected visual consistency. Also, the horse appears to stand on water rather than solid ground, which is unrealistic. These signs of incomplete objects and missing reflections strongly indicate AI generation. \\
Para-2 & The left hind leg of the horse is conspicuously missing, resulting in an awkward stance. Additionally, the shape of the ears and face presents inconsistencies that imply digital manipulation. The expected reflection on the water is absent, disrupting the visual coherence. Furthermore, the horse appears to be walking on water, which is implausible. These indicators of incomplete elements and absent reflections strongly suggest AI generation. \\
Para-3 & The horse's left hind leg is clearly missing, leading to an unnatural posture. There are also discrepancies in the shape of the ears and face that hint at digital alterations. The reflection on the water is lacking, which undermines the visual consistency. Moreover, the horse seems to be standing on water instead of solid ground, which is unrealistic. These signs of incomplete objects and missing reflections strongly indicate AI generation. \\
Para-4 & The absence of the horse's left hind leg is striking, resulting in an unnatural pose. The ears and face exhibit irregular shapes that suggest they have been digitally altered. The reflection on the water is missing where it should be, disrupting the expected visual harmony. Additionally, the horse appears to be walking on water, which is not realistic. These indicators of incomplete objects and absent reflections strongly imply AI generation. \\
\bottomrule
\end{tabularx}
\caption{Comparison of a fake image and the corresponding paraphrased and Gold-standard reasoning produced by our bootstrapping process. }
\end{table*}

\clearpage
\section{Training hyperparameters}
\label{sec:training_hyperparameters}
We ran training on two AMD-MI210 GPUs, with 64 GB of memory each. DeepfakeJudge-7B took 4 hours to train, while DeepfakeJudge-3B took 2.5 hours. Table \ref{tab:hyperparameters}
shows the exact hyperparameters used while training our DeepfakeJudge models.
\begin{table}[h]
\begin{center}
    \begin{adjustbox}{max width=\linewidth}
    \begin{tabular}{l|l}
    \toprule
    \scriptsize
    \textbf{Hyperparameter} & \textbf{Value} \\
    \midrule
    \multicolumn{2}{c}{\textbf{Training}}\\
    Learning rate & 1e-6 \\
    Train type & Lora \\
    Lora Rank & 32 \\
    Lora Alpha & 64 \\
    Batch size (per device) & 16 \\
    Gradient accumulation steps & 1 \\
    Num. train epochs & 2 \\
    Warmup ratio & 0.05 \\
    Weight decay & 0.0 \\
    Precision & bfloat16 \\
    Max sequence length & 128{,}000 \\
    tune mm vision & False \\
    tune mm mlp & True \\
    tune mm llm & True \\
    \midrule
    \multicolumn{2}{c}{\textbf{Evaluation}}\\
    max new tokens & 2048 \\
    temperature & 0.0 \\
    \bottomrule
    \end{tabular}
    \end{adjustbox}
\end{center}
\caption{Training and evaluation hyperparameters.}
\label{tab:hyperparameters}
\end{table}


\section{Inter-annotator Statistics for DeepfakeJudge-Meta-Human dataset.}
Table \ref{tab:human_annotation_summary} shows the pairwise and pointwise evaluation scores while preparing the DeepfakeJudge-Human dataset. 
\label{sec:agreement_stats}
\begin{table}
\begin{center}
    \begin{adjustbox}{max width=\linewidth}
    \begin{tabular}{l|l}
    \toprule
    \tiny
    \textbf{Metric} & \textbf{Value} \\
    \midrule
    \multicolumn{2}{c}{\textbf{Pairwise Evaluation}} \\
    Samples used & 100 per annotator \\
    Annotator 1 accuracy (vs. GT) & 0.91 \\
    Annotator 2 accuracy (vs. GT) & 0.95 \\
    Raw agreement (Annotator 1 \& 2) & 0.90 \\
    Cohen’s $\kappa$ & 0.80 \\
    Both correct & 88 \\
    Both wrong & 2 \\
    Disagree, one correct & 10 \\
    \midrule
    \multicolumn{2}{c}{\textbf{Pointwise Evaluation}} \\
    Samples used & 100 per annotator \\
    Annotator 1 MSE / RMSE & 0.11 / 0.33 \\
    Annotator 2 MSE / RMSE & 0.27 / 0.52 \\
    Inter-annotator MSE / RMSE & 0.39 / 0.62 \\
    Annotator 1 exact match rate & 0.89 \\
    Annotator 2 exact match rate & 0.73 \\
    Pearson $r$ (Annotator 1 / 2 / Overlap) & 0.97 / 0.93 / 0.90 \\
    Spearman $\rho$ (Annotator 1 / 2 / Overlap) & 0.97 / 0.93 / 0.90 \\
    Both exact correct & 65 \\
    One correct & 31 \\
    Both wrong (same / different) & 2 / 2 \\
    \bottomrule
    \end{tabular}
    \end{adjustbox}
\end{center}
\caption{Summary of human annotation reliability across pairwise and pointwise evaluations using annotators 1 and 2. Values are computed against automatically-derived ground truth.}
\label{tab:human_annotation_summary}
\end{table}

\clearpage
\label{section: all_prompts}
\begin{figure*}[t]
  \centering
  \vspace{0.0em}
  \textbf{\LARGE Prompts}
  \vspace{2em}
  
  \vspace{0.5em}
  \textbf{\large Prompt Template for Degraded Reasoning Generation}
  \vspace{0.5em}

  \begin{tcolorbox}[
      colback=codegray,
      colframe=codeborder,
      arc=3pt,
      boxrule=0.6pt,
      left=2mm, right=2mm, top=1mm, bottom=1mm
  ]
  \lstset{style=promptlisting}
\begin{lstlisting}
You are provided with an image, the ground truth classification (real/fake/edited), a description about the reasoning rated 5 (perfectly accurate, highest quality), a JSON list showing the human annotation on the image (raised flags, bounding boxes normalized to 1-1000, and referring expressions), and a flag taxonomy of human-familiar forensic cues used by annotators. Your task is to generate four additional responses that simulate progressively lower-quality reasoning outputs for the same image. Each generated response should correspond to a quality rating from 4 to 1, where Rating 5 is the provided gold standard and Ratings 4 through 1 represent decreasing quality. As the rating decreases, the responses should reflect increasing levels of degradation (for example, wrong classification, hallucinations, omissions, irrelevant details, logical inconsistencies, misused or missed flags, grammatical issues, or wording issues). Only do wrong classification for very low ratings. All generated responses must be similar in length to the gold standard and maintain the expected task format (for example, short answer plus explanation). Keep in mind that for real images there is no human annotation; the gold standard is based on the absence of certain random flags. When misclassifying, misclassify between the three classes randomly (real <-> fake <-> edited). For edited images, the label also contains the bounding box for the fake region, so use it when creating lower-rating reasoning outputs when necessary. Do not simply truncate the gold response. Use the image and human annotation to ground correctness; vary style and wording naturally. Make sure that the answers are degraded based on human perception. The higher rated answers are more useful to humans for understanding the outputs.

Task:
Generate four degraded responses corresponding to quality ratings 4 through 1, based on the gold standard response (Rating 5). Follow the rating guidelines strictly.

Rating Guidelines:
- Rating 4 - Mostly accurate with minor issues. Preserves key meaning; small lapses in detail or precision; flags mostly correct; clear wording, random wrong flag added/removed. 
- Rating 3 - Partially correct. Conveys the general idea but with noticeable errors, omissions, or misinterpretations; some generic phrasing; flags partly correct or incomplete; clarity or helpfulness mixed; unclear wording; sometimes underconfident tone but correct classification. 
- Rating 2 - Poor alignment (may misclassify). Serious flaws; weak grounding; mixes valid and invalid details; flags often wrong or irrelevant; vague or hedged language; sometimes underconfident tone but wrong classification. 
- Rating 1 - Unrelated or incorrect (often misclassifies). Fails to reflect the image or instruction; hallucinated content; misleading or nonsensical flags; confusing wording; sometimes high-confidence tone but wrong classification.

Output Format:
Return a JSON object with exactly the keys "rating_4", "rating_3", "rating_2", and "rating_1". Each value must be only the generated response text for that quality rating (no extra fields).

Flag Taxonomy:
{Flag descriptions}
EXAMPLE:
Image: 
See the first attached image.
Label:
real
Human Annotation:
{}

Input:
Image:
{Image}
Label (contains the edited region for edited class):
{Label}
Human Annotation (only for fake and edited):
{Human annotation}
Gold Standard Response (Rating 5):
{gold_standard_response}

Output (according to the JSON schema):
{
"rating_4": "[Generated response]",
"rating_3": "[Generated response]",
"rating_2": "[Generated response]",
"rating_1": "[Generated response]"
}
"""

\end{lstlisting}
  \end{tcolorbox}
  \label{fig: prompts}
\end{figure*}

\begin{figure*}[htbp]
  \centering
  \vspace{0.5em}
  \textbf{\large Prompt Template for Evaluator model.}
  \vspace{0.5em}

  \begin{tcolorbox}[
      colback=codegray,
      colframe=codeborder,
      arc=3pt,
      boxrule=0.6pt,
      left=2mm, right=2mm, top=1mm, bottom=1mm
  ]
  \lstset{style=promptlisting}
\begin{lstlisting}
You are provided with an image, the ground truth classification (real / fake / edited), a gold-standard reasoning rated 5 (perfectly accurate, highest quality), a JSON list of human annotations for the image (raised flags, any bounding boxes normalized to 1-1000, and referring expressions), a flag taxonomy of human-familiar forensic cues, and a set of candidate responses (each typically containing a short answer plus an explanation).

Your task is to evaluate each candidate response and assign an evaluation rating from 1 to 4. While the gold-standard response (Rating 5) is a reference for what ideal reasoning looks like, your evaluation should primarily judge each candidate on how well it fulfills the instruction for this image, grounded in the image content, label, and human annotations. Focus your reasoning on what is incorrect, missing, misleading, or poorly justified in the candidate response itself.

Special notes and constraints:

- For real images, the human annotation is typically empty; the gold reasoning reflects the absence of artifact flags. Penalize candidates that invent edited regions, hallucinate artifacts, or misuse flags without evidence.
- For fake (fully generated/synthetic) images, human annotations may include raised flags but no edited-region bounding box. Evaluate whether the candidate's flags match plausible generative artifacts and whether claims are grounded in visible cues.
- For edited images, the label includes the edited region bounding box(es). Evaluate whether candidates correctly identify the manipulation and localize it (if they attempt localization). If a candidate provides a box or a referring expression, check it against the provided edited region(s) for plausibility (qualitatively; coarse overlap is acceptable). Missing or obviously wrong localization is a major error.
- Do not reward or penalize writing style alone; judge clarity insofar as it affects usefulness and correctness. Avoid over-reliance on the gold wording, evaluate the content and grounding.
- Treat the flag taxonomy below as the authoritative list of cues. Using non-existent flags, or misnaming/misapplying flags, should be penalized proportionally.

---

Evaluation Steps

1. Read the instruction to understand the expected output format (e.g., "short answer + explanation") and task focus (classification, localization, reasoning).
2. Use the image + label + human annotations (and the edited-region box if applicable) as the factual basis.
3. Refer to the gold-standard response (Rating 5) only as a reference for correctness and completeness, not as wording to match.
4. For each candidate response, identify:
   - Classification correctness (does the short answer match the label?).
   - Grounding quality (are cited cues visible and consistent with the taxonomy?).
   - Proper use (or misuse) of flags, localization, sizes/coordinates, and referring expressions.
   - Hallucinations, contradictions, irrelevancies, omissions, or logical gaps.
5. Assign a rating from 1-4 to each candidate.
6. Write a brief rationale for each rating, pointing to specific strengths/weaknesses (e.g., correct flags but missed localization; misclassification; hallucinated halos; misapplied perspective).

---

Rating Guidelines (Evaluation)

Rating Guidelines:
- Rating 4 - Mostly accurate with minor issues. Preserves key meaning; small lapses in detail or precision; flags mostly correct; clear wording, random wrong flag added/removed. 
- Rating 3 - Partially correct. Conveys the general idea but with noticeable errors, omissions, or misinterpretations; some generic phrasing; flags partly correct or incomplete; clarity or helpfulness mixed; unclear wording; sometimes underconfident tone but correct classification. 
- Rating 2 - Poor alignment (may misclassify). Serious flaws; weak grounding; mixes valid and invalid details; flags often wrong or irrelevant; vague or hedged language; sometimes underconfident tone but wrong classification. 
- Rating 1 - Unrelated or incorrect (often misclassifies). Fails to reflect the image or instruction; hallucinated content; misleading or nonsensical flags; confusing wording; sometimes high-confidence tone but wrong classification.

---

Output Format

Return only a JSON object whose keys exactly mirror the candidate identifiers provided in Generated Responses (e.g., "candidate-1", "candidate-2",...).  
Each value must be a JSON object with exactly these keys:

- "rating": an integer in {1, 2, 3, 4}  
- "rationale": a concise string explaining the rating (no more than a few sentences)

Do not include any additional keys or text outside the JSON object.

---

Flag Taxonomy (Reference for Evaluation)
{Flags}

EXAMPLE  
Image: See the first attached image.  

Label: real  

Human Annotation: {}

Output Format:
{"candidate_1": {"rating": <1-4>, "rationale": "<concise explanation>"}, "...": {...}}

Input:

Image:
See attached image.
Label:
{label_text}
Human Annotation:
{human_annotation}
Gold Standard Response (Rating 5):
{gold_standard_response}
Generated Responses:
{generated_responses}
Output (according to the JSON schema):
{output_format}

\end{lstlisting}
  \end{tcolorbox}
  \label{fig:first_prompt}
\end{figure*}

\begin{figure*}[htbp]
  \centering
  \vspace{0.5em}
  \textbf{\large Prompt Template for Regeneration.}
  \vspace{0.5em}

  \begin{tcolorbox}[
      colback=codegray,
      colframe=codeborder,
      arc=3pt,
      boxrule=0.6pt,
      left=2mm, right=2mm, top=1mm, bottom=1mm
  ]
  \lstset{style=promptlisting}
\begin{lstlisting}
You are provided with an image, an instruction, the ground-truth classification (real / fake / edited), a gold standard response rated 5 (perfectly accurate, highest quality), a JSON list of human annotations for the image (raised flags, any edited-region bounding boxes normalized to 1-1000, and referring expressions), a flag taxonomy of human-familiar forensic cues, and a set of generated responses, each intended to match a specific quality rating from 4 to 1. However, some generated responses were evaluated and found to deviate from their intended quality levels.

Task:
Your task is to regenerate revised responses only for those entries where the absolute difference between the intended rating and the evaluation rating is greater than zero (i.e., |intended rating - eval rating| > 0). For each such entry, produce a revised response that strictly conforms to the intended quality rating, based on the definitions in the rating guidelines below. Use the image and human annotations (and edited-region box if applicable) as the factual grounding for determining what content is valid. Adjust the response so that its evaluation rating would now exactly match the intended rating.
- If the eval rating is higher than the intended rating, degrade the response by introducing errors such as mild hallucinations, factual distortions, omissions, vagueness, or grammar issues (consistent with the target rating).
- If the eval rating is lower than the intended rating, improve the response by clarifying cues, reducing errors, restoring key context from the annotations/taxonomy, and aligning tightly with the label and visible evidence.

Rating Guidelines:
- Rating 4 - Mostly accurate with minor issues. Preserves key meaning; small lapses in detail or precision; flags mostly correct; clear wording.
- Rating 3 - Partially correct. Conveys the general idea but with noticeable errors, omissions, or misinterpretations; some generic phrasing; flags partly correct or incomplete; clarity or helpfulness mixed; underconfident tone but correct classification.
- Rating 2 - Poor alignment (may misclassify). Serious flaws; weak grounding; mixes valid and invalid details; flags often wrong or irrelevant; vague or hedged language; underconfident tone but wrong classification.
- Rating 1 - Unrelated or incorrect (often misclassifies). Fails to reflect the image or instruction; hallucinated content; misleading or nonsensical flags; confusing wording, high-confidence tone but wrong classification.

Flag Taxonomy :
{Flags description}
Example: 
Image: See the first attached image.  

Label: real  

Human Annotation: {}
{}


Output Format:
Your final output must be a valid JSON object. For each entry where |intended rating - eval rating| > 0, include a key "rating {i}" and update the value with a newly revised response that adheres precisely to the intended quality rating. If the eval rating is equal to the intended rating, do not modify or include that entry. Only output the revised responses.

Input:

Image:
{image}
Label (real / fake / edited):
{label}
Human Annotation (JSON with raised flags, boxes 1-1000, referring expressions):
{human_annotation}
Gold Standard Response (Rating 5):
{gold_standard_response}
Feedback Data (JSON):
{feedback_data}

Output (according to the JSON schema):
{output_format}
\end{lstlisting}
  \end{tcolorbox}
\end{figure*}

\begin{figure*}[htbp]
  \centering
  \vspace{0.5em}
  \textbf{\large Prompt Template for Rephrasing.}
  \vspace{0.5em}

  \begin{tcolorbox}[
      colback=codegray,
      colframe=codeborder,
      arc=3pt,
      boxrule=0.6pt,
      left=2mm, right=2mm, top=1mm, bottom=1mm
  ]
  \lstset{style=promptlisting}
\begin{lstlisting}
You are given context about an image-authenticity reasoning task.

Instruction:
{instruction}

Label: {label}
Human Annotation: {human_annotation}

Original response (to paraphrase):
{response_text}

Task:
Generate four paraphrases that preserve every factual claim, label decision, flag usage, bounding box reference, and XML-style tag structure (e.g., <reasoning>...</reasoning>, <answer>...</answer>). Each paraphrase should be similar in length to the original but use distinct wording. Do NOT introduce new information, remove evidence, or change the classification. Retain the same overall format and tags.

Output only a JSON object: {{"paraphrase_1": "...", "paraphrase_2": "...", "paraphrase_3": "...", "paraphrase_4": "..."}}"""

\end{lstlisting}
  \end{tcolorbox}
\end{figure*}

\begin{figure*}[htbp]
  \centering
  \vspace{0.5em}
  \textbf{\large Prompt Template for Rephrasing.}
  \vspace{0.5em}

  \begin{tcolorbox}[
      colback=codegray,
      colframe=codeborder,
      arc=3pt,
      boxrule=0.6pt,
      left=2mm, right=2mm, top=1mm, bottom=1mm
  ]
  \lstset{style=promptlisting}
\begin{lstlisting}
You are given context about an image-authenticity reasoning task.

Instruction:
{instruction}

Label: {label}
Human Annotation: {human_annotation}

Original response (to paraphrase):
{response_text}

Task:
Generate four paraphrases that preserve every factual claim, label decision, flag usage, bounding box reference, and XML-style tag structure (e.g., <reasoning>...</reasoning>, <answer>...</answer>). Each paraphrase should be similar in length to the original but use distinct wording. Do NOT introduce new information, remove evidence, or change the classification. Retain the same overall format and tags.

Output only a JSON object: {{"paraphrase_1": "...", "paraphrase_2": "...", "paraphrase_3": "...", "paraphrase_4": "..."}}"""

\end{lstlisting}
  \end{tcolorbox}
\end{figure*}

\begin{figure*}[htbp]
  \centering
  \vspace{0.5em}
  \textbf{\large Prompt Template for pointwise model evaluation.}
  \vspace{0.5em}

  \begin{tcolorbox}[
      colback=codegray,
      colframe=codeborder,
      arc=3pt,
      boxrule=0.6pt,
      left=2mm, right=2mm, top=1mm, bottom=1mm
  ]
  \lstset{style=promptlisting}
\begin{lstlisting}
You are provided with a image, a corresponding ground truth class ( real/fake/edited), and a response generated by a model. The response consists of the model's final prediction, along with a rationale describe how the model reached that decision.
Your task is to evaluate the quality of the response, considering how well it explains the classification behind its prediction, based on
the content of the image and the ground truth. This is a holistic judgment and should be based on the overall correctness, relevance, completeness, and
grounding of the response.
Task:
For each response:
- Assess how well it is describes the reasoning behind the prediction, and how well it is grounded in the image.
- Consider whether the response is accurate, relevant, complete, and grounded in the image. For edited images, also use the bounding box information for rating the responses.
- Provide a brief rationale explaining the overall quality and alignment of the response inside <reasoning></reasoning>.
- Output a score from 1 (worst) to 5 (best) indicating the overall quality inside <score></score>.

Rating Guidelines:
- Rating 5 - Fully accurate, complete, and well-grounded. The response precisely explain the reason behind the correct classification with no notable errors or omissions.
- Rating 4 - Mostly accurate with minor issues. Preserves key meaning; small lapses in detail or precision; flags mostly correct; clear wording, random wrong flag added/removed. 
- Rating 3 - Partially correct. Conveys the general idea but with noticeable errors, omissions, or misinterpretations; some generic phrasing; flags partly correct or incomplete; clarity or helpfulness mixed; unclear wording; sometimes underconfident tone but correct classification. 
- Rating 2 - Poor alignment. Serious flaws; weak grounding; mixes valid and invalid details; flags often wrong or irrelevant; vague or hedged language; sometimes underconfident tone but wrong classification. 
- Rating 1 - Unrelated or incorrect. Fails to reflect the image or instruction; hallucinated content; misleading or nonsensical flags; confusing wording; sometimes high-confidence tone but wrong classification.

Input:
Image: Check the attached image

Ground Truth Label: {label}

Candidate response:
{candidate_response}
Output (Strict Format):
<reasoning>{{your reasoning and explanation for the rating}}</reasoning>
<score>{{integer score from 1 to 5}}</score>
\end{lstlisting}
  \end{tcolorbox}
\end{figure*}

\begin{figure*}[htbp]
  \centering
  \vspace{0.5em}
  \textbf{\large Prompt Template for pairwise model evaluation.}
  \vspace{0.5em}

  \begin{tcolorbox}[
      colback=codegray,
      colframe=codeborder,
      arc=3pt,
      boxrule=0.6pt,
      left=2mm, right=2mm, top=1mm, bottom=1mm
  ]
  \lstset{style=promptlisting}
\begin{lstlisting}
You are provided with an image, a corresponding ground-truth class (real/fake/edited), and two responses generated by different models. Each response consists of the model's final prediction plus a rationale.
Your task is to compare the two responses and decide which one better aligns with the ground truth classification, based on the content of the image. Make a holistic judgment that considers correctness, relevance, completeness, and grounding.
Task:
For the given pair of responses:
- Judge which response is better reasoning for the classification
- Consider whether each response is accurate, relevant, complete, and grounded in the video.
- Output only A or B, wrapped strictly inside <answer></answer> tags.

Evaluation Guidelines:
- Accuracy: Prefer responses that are factually correct and consistent with the image, and have the correct classification
- Relevance: Prefer responses that directly mention the reasoning behind the classification
- Completeness: Prefer responses that capture all key aspects needed for a full grounded answer, containg the exact issues/non-issues with the image.
- Grounding: Prefer responses clearly supported by the image, avoiding hallucinations.
- If one response contains hallucinations, irrelevant content, or omissions, prefer the other.
- If both responses are strong, choose the one that is more precise and detailed.
- If both responses are weak, choose the one that is less flawed.

Ground Truth Label: {label}
Input:
Response A:
{response_a}

Response B:
{response_b}

Output (Strict Format):
<answer>{{A_or_B}}</answer>
\end{lstlisting}
  \end{tcolorbox}
\end{figure*}

\begin{figure*}[htbp]
  \centering
  \vspace{0.5em}
  \textbf{\large Prompt Template for generating gold standard rating from human-annotated fake data.}
  \vspace{0.5em}

  \begin{tcolorbox}[
      colback=codegray,
      colframe=codeborder,
      arc=3pt,
      boxrule=0.6pt,
      left=2mm, right=2mm, top=1mm, bottom=1mm
  ]
  \lstset{style=promptlisting}
\begin{lstlisting}
You will receive: (1) an image and (2) a JSON-like string named "flags" listing issues. 
Each entry includes "bboxes": [{"coordinates":[x1,y1,x2,y2], "ref_exp":"..."}] with coordinates normalized 1-1000; entries may include "timestamp" (ignore it).

USE OF FLAGS
Treat the entries as hints about WHAT and WHY specific regions look synthetic. 
Never mention JSON, flags, coordinates, boxes, or timestamps in your writing. Write as if you are simply looking at the image and grounding your observations in visible evidence consistent with the hints.
Convert some regions into **coarse spatial phrases** (upper-left, upper-center, upper-right, mid-left, center, mid-right, lower-left, lower-center, lower-right) **only when the SAME anomaly occurs multiple times** and needs differentiation; otherwise, omit location phrases and refer to the objects directly.
Ignore the referring expression if it does not entail a flaw in the image, but rather contain instructions about things that might have been done to the image
OUTPUT FORMAT (exactly two lines, nothing else):
<think>
Write 4-6 sentences (about 70-120 words) describing concrete observations. Integrate and paraphrase the hinted issues naturally (e.g., "skin looks unnaturally smooth," "hands are brighter than nearby skin," "faces are distorted"). Use at least FOUR distinct cues overall: one texture/material cue, one lighting/color cue, one geometry/structure cue, and one boundary/depth cue. Conclude that these problems indicate the image is AI-generated.
</think>
<answer>fake</answer>

RULES
Neutral, observational tone; do not identify real people.
Vary phrasing while staying faithful to the visible evidence and the hinted issues.
\end{lstlisting}
  \end{tcolorbox}
\end{figure*}

\begin{figure*}[htbp]
  \centering
  \vspace{0.5em}
  \textbf{\large Prompt Template for generating gold standard rating from real data.}
  \vspace{0.5em}

  \begin{tcolorbox}[
      colback=codegray,
      colframe=codeborder,
      arc=3pt,
      boxrule=0.6pt,
      left=2mm, right=2mm, top=1mm, bottom=1mm
  ]
  \lstset{style=promptlisting}
\begin{lstlisting}

You are an image-authenticity checker. For this task the ground-truth label is Real. Keep this private. Your job is to produce a short, observation-based justification that naturally supports a Real verdict.

OUTPUT FORMAT (exactly two lines, nothing else):
<think>[4-7 sentences; 70-130 words of observations]</think>
<answer>Real

DELIVERABLE RULES
Base the <think> text only on what is visible in the image. Refer to concrete objects (e.g., "plate," "cherries," "helmet," "gravel," "glasses").  
Use at least FOUR distinct visual cues, including at least:
  - one photometric cue (shadows, lighting, or color cast),  
  - one geometric cue (scale or perspective/line convergence),  
  - one boundary/depth cue (contact/support, overlap, focus/blur, or edges/halo).  
Write in present tense, neutral and concise; avoid purple prose, probabilities, or self-references.  
Do NOT mention any "checklist," "flags," or that you were given the label.  
Do NOT identify real people; describing non-identifying attributes (e.g., clothing, pose) is fine.  
No bullet lists or newlines inside <think>; use sentences only.  
The second line must read exactly: <answer>Real</answer>

CUE LIBRARY (use, but do not name explicitly)
1) Shadows - Directions align; dark contact at touchpoints; edges soften with distance.  
2) Lighting Match - Same light side/contrast across scene.  
3) Color Cast Consistency - Whites/neutrals share similar warm/cool tint.  
4) Relative Size / Scale - Familiar objects are plausibly sized and diminish with distance.  
5) Perspective / Line Convergence - Parallel edges aim to consistent vanishing point(s).  
6) Front-Back Overlap - Near objects cleanly occlude far ones; no halos/see-through.  
7) Contact & Support - Objects rest firmly; soft materials compress; no "floating."  
8) Focus / Blur with Depth - Sharpness changes smoothly with distance; neighbors at same depth share sharpness.  
9) Reflections & Transparency - Reflections show correct content/angle; glass shows some reflection and some see-through.  
10) Material Shine - Shiny = crisp highlights; matte = broad/soft highlights.  
11) Texture / Pattern Follow-Through - Stripes/grain bend and scale with form/folds.  
12) Text & Small Details - Letters legible; tiny parts coherent.  
13) Object Completeness & Counts - Obvious parts present and counted correctly.  
14) Edges & Boundaries - Edge sharpness matches surroundings; no glow, color fringing, or scissor-cut look.  
15) Other - Any additional visual consistency that supports realism.

QUALITY NOTES
Vary which cues you emphasize across images to avoid repetitive language.  
Always tie observations to specific objects and spatial relations rather than abstract claims.

Remember: exactly two lines of output, with the second line fixed to <answer>Real</answer>
\end{lstlisting}
  \end{tcolorbox}
\end{figure*}

\clearpage

\section{Ethics Statement}
Our work raises important ethical considerations, particularly due to the use of visual data and human participation in deepfake research. The dataset and human studies were conducted following institutional guidelines and community standards to ensure responsible and transparent research. Below, we outline all ethical considerations related to this work.

\paragraph{Use of Public and Generated Visual Data:}
All images used in DeepfakeJudge originate from publicly available datasets or from synthetically generated sources. The real subset is derived from Open-Images V7 \cite{OpenImages} from Google, used under its research license with appropriate attribution. The synthetic and edited subsets were created using text-to-image and image-editing models, including Gemini \cite{comanici2025gemini25pushingfrontier}, SeedDream \cite{seedream2025seedream40nextgenerationmultimodal}, Flux-Kontext \cite{labs2025flux1kontextflowmatching}, and Qwen-Edit \cite{wu2025qwenimagetechnicalreport}, solely for academic and non-commercial research. Each generated or edited image underwent linguistic, semantic, and NSFW filtering, followed by manual verification to remove inappropriate, offensive, or personally identifiable content. This ensures that all data are ethically sourced and safe for research use. The collection and use of such material are consistent with established practices followed in peer-reviewed benchmarks such as FaceForensics++ \cite{rössler2019faceforensicslearningdetectmanipulated} and DFDC \cite{dolhansky2020deepfakedetectionchallengedfdc}, which also use public or generated content for research under fair-use provisions.

\paragraph{Human Annotation:}
The first phase of human annotation involved six trained annotators who labeled reasoning cues, bounding boxes, and referring expressions across both in-distribution and out-of-distribution subsets. All annotators were over 18 years of age, affiliated with university research groups, and completed a shared pilot phase to ensure consistency. Inter-annotator agreement reached substantial alignment (Cohen’s $\kappa=0.71$). All annotators were compensated for their time and effort. No personal information was collected, and all materials were screened to exclude NSFW or sensitive content. Participation was voluntary, and all annotations were used exclusively for research purposes.
A second phase of annotation was conducted to create the DeepfakeJudge-Meta-Human dataset. Two expert annotators independently evaluated reasoning quality and pairwise reasoning preferences. Both annotators were over 18 years old, recruited through academic networks, and compensated for their work. We evaluated overlapping subsets to ensure reliability, resulting in high inter-annotator agreement (Cohen’s $\kappa=0.80$ for pairwise and MSE = 0.39 for pointwise evaluation). All participation was voluntary, and no personal data were collected.

\paragraph{User Study:}
The user study evaluating reasoning preferences followed institutional IRB guidelines. Ten adult participants were recruited from research groups at the authors’ affiliated universities. Participation was voluntary, and participants were compensated. Before participation, each subject received an explanatory statement outlining the study objectives and procedures, along with a few examples detailing the procedure, and informed consent was obtained through the study form. All materials used in the study were screened to exclude NSFW or distressing content. No personal information or identifying data were recorded, and all responses were anonymized.

\paragraph{Access and Use Policy:}
All datasets, models, and code will be released for academic, non-commercial use only, under a strict End-User License Agreement (EULA). Access will be granted only to verified academic researchers under the following conditions:
(1) Use is limited to academic, educational, and not-for-profit research.
(2) Each institution accepts responsibility for its authorized users.
(3) Redistribution, modification, or misuse of the data is prohibited.
(4) Access may be revoked or modified by licensors at any time.
(5) Any use that could cause embarrassment, harm, or distress to subjects is strictly forbidden.

This controlled-access policy aligns with established community practices, such as Faceforensics++ \cite{rössler2019faceforensicslearningdetectmanipulated} and DFDC \cite{dolhansky2020deepfakedetectionchallengedfdc}, ensuring that research on deepfake detection and reasoning remains ethical, traceable, and responsibly conducted. All stages of this work, including data creation, annotation, and user evaluation, adhered to institutional and community standards for human research ethics.

\section{Effect of Paraphrasing}
\begin{table}[htbp]
\centering
\caption{Performance of DeepfakeJudge-3B on the DeepfakeJudge-Meta dataset with and without paraphrasing.}
\small
\begin{tabular}{lccccc}
\toprule
{Model} & {Para} & {RMSE\,($\downarrow$)} &{MSE\,($\downarrow$)} & {s\,($\uparrow$)} & {p\,($\uparrow$)} \\
\midrule
DeepfakeJudge-3B & \ding{51} & 0.69 & 0.48 & 0.92 & 0.92 \\
DeepfakeJudge-3B & \ding{55} & 1.39 & 1.93 & 0.60 & 0.62 \\
\bottomrule
\label{tab:para_pointwise}
\end{tabular}
\end{table}
\begin{table}[htbp]
\centering
\caption{Pairwise accuracy of DeepfakeJudge-3B on the DeepfakeJudge-Meta dataset with and without paraphrasing.}
\small
\begin{tabular}{lcc}
\toprule
{Model} & {Paraphrasing} & {DFJM\,(\%)} \\
\midrule
DeepfakeJudge-3B & \ding{51} & 94.4 \\
DeepfakeJudge-3B & \ding{55} & 89.1 \\
\bottomrule
\label{tab:para_pairwise}
\end{tabular}
\end{table}

Tables \ref{tab:para_pointwise}, \ref{tab:para_pairwise} show the relative performance of DeepfakeJudge-3B model, trained with and without paraphrasing. These results signify the importance of paraphrasing to avoid overfitting on linguistic styles, thereby forcing the model to focus on semantic information instead. 

\section{Out-of-Distribution Generalization of DeepfakeJudge Model}

\begin{table*}[h!]
\centering
\renewcommand{\arraystretch}{1.4}
\setlength{\tabcolsep}{5pt}
\caption{Rationales and scores generated by DeepfakeJudge-3B on the ground truth reasonings of the test-set of DD-VQA. DeepfakeJudge-3B correctly identifies where the reasoning is ungrounded and superficial, and gives a rating accordingly.}
\label{tab: examples_ddvqa}
\begin{tabularx}{\textwidth}{
|>{\centering\arraybackslash}m{3cm}|
>{\arraybackslash}m{7cm}|
>{\arraybackslash}m{5cm}|
>{\centering\arraybackslash}m{1cm}|}
\hline
\textbf{Image} & \textbf{Ground Truth Reasoning (from DDVQA)} & \textbf{Evaluation Response} & \textbf{Score} \\
\hline
\includegraphics[width=3cm]{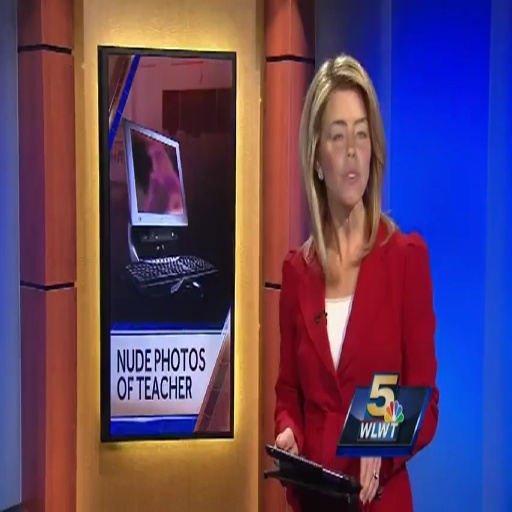} &
The image looks fake. It is an image with manipulated face regions. The person's mouth looks a bit fake because overly small mouth underage individuals. The person's skin looks a bit fake because the skin is overly smooth and lacks of details, such as skin texture the person has mismatched bangs. &
This response correctly identifies the manipulated face regions and mentions the smooth skin, which aligns with the ground truth label. However, it incorrectly suggests the person is underage, which is not supported by the image. &
3 \\
\hline

\includegraphics[width=3cm]{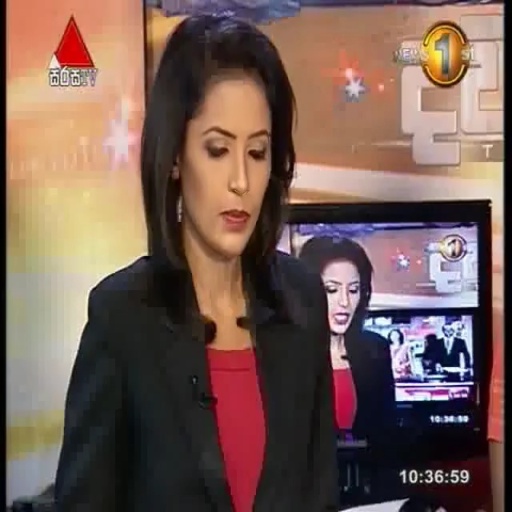} &
The image looks fake. It looks like a computer-generated artist styled image for animation, carton or films. The person's eyes look very fake because the eyes look unnaturally asymmetric. The person has mismatched bangs. The person's nose looks very fake because the nose lookes curved. &
This response correctly identifies the image as fake, noting asymmetry and mismatched features. However, it lacks specificity about the style and suggests the person is animated, which is not entirely accurate. &
3 \\
\hline

\includegraphics[width=3cm]{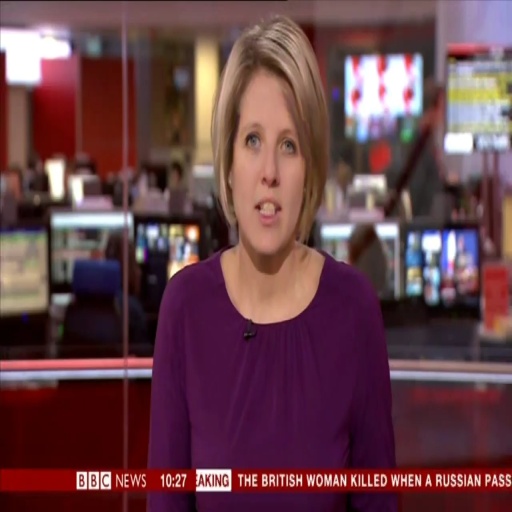} &
The image looks fake. It looks like an ai-generated image although with human appearance but with unrealistic texture or lighting. The person's eyes look a bit fake because the eyes look unnaturally asymmetric. The eyes looks too rigid and lifeless to convey expresssion. The person has mismatched bangs. Mismatched mustache. The lighting in the image is inconsistent, unrealistic shadows. The person's nose looks very fake because the nose lookes curved. The nose is placed at the wrong place on the face. &
Fully accurate, complete, and well-grounded. The response precisely explain the reason behind the correct classification with no notable errors or omissions. &
5 \\
\hline

\includegraphics[width=3cm]{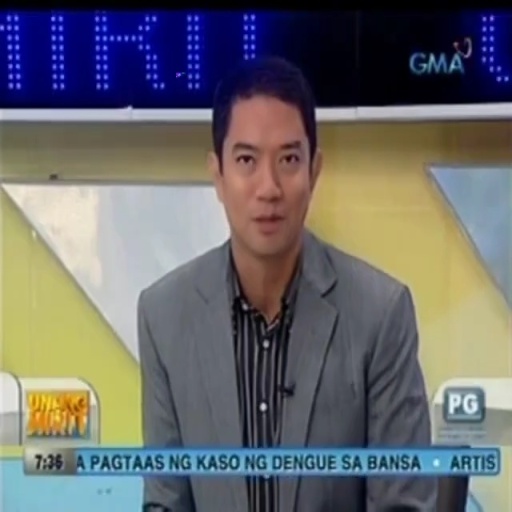} &
The image looks fake. It is an image with manipulated face regions. The person's skin looks very fake because the skin is overly smooth and lacks of details, such as skin texture there are stains/flaws on the person's face there are areas looking unnaturally brighter there area of looking unnaturally darker the person's nose looks a bit fake because the nose do not align properly with other facial features. The person's eyebrows look a bit fake because the eyebrows do not match the face's structure. &
This response accurately identifies the manipulation of facial features, noting the lack of natural skin texture and the unnatural brightness and darkness. However, it could have been more specific about the alignment issues with the nose and eyebrows. &
4 \\
\hline

\end{tabularx}

\end{table*}

To demonstrate that our models, DeepfakeJudge-3B and DeepfakeJudge-7B, can effectively evaluate deepfake reasoning on data outside our training pipeline, we assess them using the ground truth human-annotated reasonings from the DD-VQA dataset \cite{zhang2024commonsensereasoningdeepfake}, which contains reasonings for face-swapped images. We convert the test set to our input format, placing the reasoning within \texttt{<reasoning>...</reasoning>} and the prediction inside \texttt{<answer>...</answer>}. The model then generates a rating along with an explanatory rationale. On average, our model assigns a rating of \textbf{\textit{3.18}} across all responses. Table \ref{tab: examples_ddvqa} presents several qualitative examples of these ratings and the corresponding rationales. The rationales produced by our model are clear, informative, and well-grounded in the visual content. Moreover, these rationales provide valuable interpretability and can serve as reliable signals for large-scale automatic supervision or self-evaluation pipelines in future deepfake detection systems.


\end{document}